\documentclass{article}
\usepackage{iclr2027_conference,times}

\usepackage{amsmath,amsfonts,bm}

\def\eqref#1{equation~\ref{#1}}

\def\1{\bm{1}}

\DeclareMathAlphabet{\mathsfit}{\encodingdefault}{\sfdefault}{m}{sl}
\SetMathAlphabet{\mathsfit}{bold}{\encodingdefault}{\sfdefault}{bx}{n}

\usepackage{amsmath}
\usepackage{amssymb}
\usepackage{amsfonts}
\usepackage{booktabs}
\usepackage{array}
\usepackage[table]{xcolor}

\definecolor{verylightgray}{gray}{0.89} 
\definecolor{RPKey}{RGB}{0, 0,0}
\definecolor{RPValue}{RGB}{0, 0,0}
\newcommand{\RPKL}{%
  \ensuremath{\textcolor{RPKey}{\mathbf{K}_{L}}}}
\newcommand{\RPVL}{%
  \ensuremath{\textcolor{RPValue}{\mathbf{V}_{L}}}}

\newcommand{\RPbarKL}{%
  \ensuremath{\textcolor{RPKey}{\bar{\mathbf{K}}_{L}}}}
\newcommand{\RPbarVL}{%
  \ensuremath{\textcolor{RPValue}{\bar{\mathbf{V}}_{L}}}}

\usepackage{graphicx}
\usepackage{float}
\usepackage{multirow}
\usepackage{microtype}
\usepackage{listings}
\usepackage{url}
\usepackage{hyperref}

\definecolor{codebg}{RGB}{249,249,249}
\definecolor{coderule}{RGB}{220,220,220}
\definecolor{codecomment}{RGB}{90,120,90}
\definecolor{codekeyword}{RGB}{80,80,80}

\lstdefinestyle{rpattenpython}{
  language=Python,
  backgroundcolor=\color{codebg},
  basicstyle=\ttfamily\normalsize,
  keywordstyle=\bfseries\color{codekeyword},
  commentstyle=\itshape\color{codecomment},
  stringstyle=\ttfamily,
  numbers=none,
  frame=single,
  rulecolor=\color{coderule},
  framerule=0.4pt,
  framesep=5pt,
  showstringspaces=false,
  breaklines=true,
  columns=fullflexible,
  keepspaces=true,
  xleftmargin=0pt,
  xrightmargin=0pt,
  aboveskip=0.6em,
  belowskip=0.6em
}

\title{Representative Attention \\ For Vision Transformers}

\author{Yuntong Li, Hengxing Liu, Hainuo Wang, Mingjia Li, Xiaojie Guo\thanks{Corresponding author.} \\
College of Intelligence and Computing, Tianjin University, Tianjin 300350, China \\
\texttt{lytong@tju.edu.cn, chrisliu.jz@gmail.com, hainuo@tju.edu.cn} \\
\texttt{mingjiali@tju.edu.cn, xj.max.guo@gmail.com}}

\iclrfinalcopy
\hypersetup{
  hidelinks,
  pdftitle={Representative Attention For Vision Transformers},
  pdfauthor={Yuntong Li, Hengxing Liu, Hainuo Wang, Mingjia Li, Xiaojie Guo}
}

\begin{document}

\maketitle

\begin{abstract}
Many efficient attention methods reduce the quadratic cost of global self-attention by using predefined spatial partitions 
to construct compact proxies or restrict interactions.
However, these partitions can separate distant tokens with similar features and group neighboring tokens with dissimilar features.
This creates a mismatch between token interactions and the feature structure of visual content.
To address this mismatch, we introduce \emph{Representative Attention} (RPAttention), 
which organizes global communication through a compact set of input-dependent representatives.
A learned projection of input token features determines their contributions to each representative.
RPAttention then constructs representatives by GATHERING information from \textbf{ALL} image tokens rather than a predefined spatial subset.
The constructed representatives then INTERACT through lightweight self-attention to refine the gathered content.
They DISTRIBUTE their refined content to each spatial query using attention weights computed separately from the gathering assignments.
With $M$ representatives and $N$ spatial tokens, the complete Gather--Interact--Distribute process requires $\mathcal{O}(NM+M^2)$ token interactions.
Experiments on image classification, object detection, instance
segmentation, and semantic segmentation demonstrate strong performance
across multiple backbone families and competitive results against
recent efficient attention methods.
Runtime measurements further show favorable scaling as input
resolution increases.

\end{abstract}

\section{Introduction}

Vision Transformers (ViTs)~\citep{vit} have become a standard architecture
for visual representation learning, with self-attention enabling direct
interactions between distant tokens.
Visual tokens are typically obtained by dividing images into patches,
and natural images contain substantial spatial redundancy~\citep{mae}.
Explicit interactions between every pair of tokens may therefore be
unnecessary for capturing global visual information.
Nevertheless, standard global self-attention computes all pairwise
interactions among $N$ spatial tokens, incurring $\mathcal{O}(N^2)$
complexity.
This quadratic cost becomes increasingly expensive at higher input
resolutions, motivating more efficient attention mechanisms.

A central question is how to reduce these interactions without losing
access to global information.
Some methods restrict attention to local windows or
neighborhoods~\citep{swin,cswin,nat}, while others retain a global
communication path through compact proxies constructed using spatial
reduction, pooling, or locally constrained
associations~\citep{pvt,pvt_v2,agent_attention,stvit}.
These approaches retain spatial priors in proxy construction
or token association.
This can separate distant tokens with similar features and combine
neighboring tokens with dissimilar features.
Spatial organization is useful for preserving visual detail, but it
need not also determine how information is aggregated for global
communication.
We therefore ask:
\emph{can each representative aggregate information from the entire
image based on token features rather than predefined spatial regions?}

\begin{figure}[t]
  \centering
  \includegraphics[width=\linewidth]{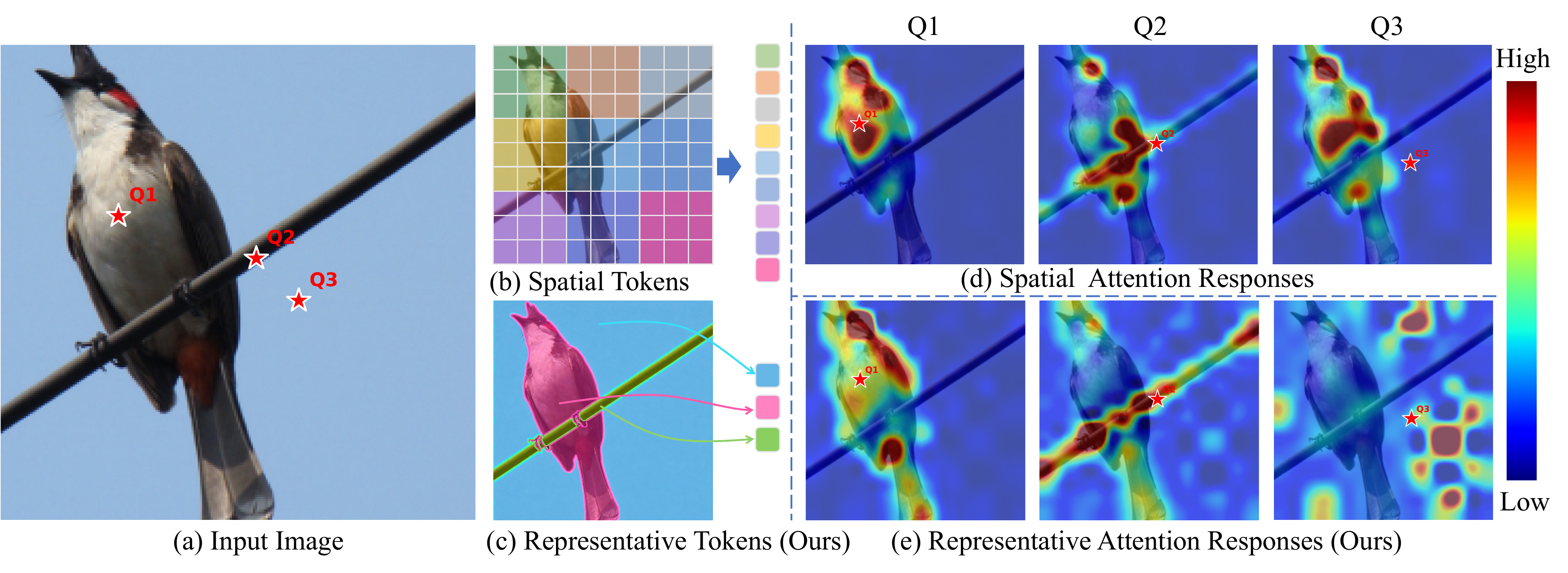}
  \caption{
    Motivation of RPAttention.
    (a) Input image with three query locations.
    (b,c) Spatial proxies follow predefined regions, whereas RPAttention
    gathers tokens globally by learned feature compatibility, allowing
    spatially distant tokens to contribute to the same representative.
    (d,e) Corresponding query-response maps.
    }
  \label{fig:motivation}
\end{figure}

We introduce \emph{Representative Attention} (RPAttention) to address
this question.
Its core design separates how tokens contribute to global
representatives from how spatial queries retrieve the refined content.
Through a \emph{Gather--Interact--Distribute} process, a learned
projection first assigns each input token soft weights over
representative anchors according to feature compatibility.
The assignments are then normalized across tokens to aggregate their
features into representatives.
Each representative can gather information from all image tokens
rather than a predefined spatial subset, and its content depends on
the current input.
The representatives interact through self-attention, after which
each spatial query retrieves their refined information through
a separately computed attention map.
As illustrated in Figure~\ref{fig:motivation}, spatially distant
tokens can contribute to the same representative, while neighboring
tokens need not share the same assignment.
RPAttention preserves the spatial token layout and has a token
interaction cost of $\mathcal{O}(NM+M^2)$ for $M$ representatives.

Global communication without fixed spatial partitions is not unique
to RPAttention.
Kernel methods aggregate features into shared
statistics~\citep{linear_attention,performer,flatten_transformer,polaformer},
while Slot Attention, Perceiver IO, and BiXT demonstrate the value
of learned latent representations~\citep{slot_attention,perceiver_io,bixt}.
Our contribution is a specific attention replacement that combines
global competitive gathering, interaction among representatives,
and independent query-conditioned distribution within standard
vision backbones.
We evaluate RPAttention across DeiT, PVT, PVTv2, and Swin on
classification, detection, and segmentation tasks, together with
analyses of representative behavior and practical efficiency.
Our contributions are threefold:
\begin{itemize}
    \item We propose organizing global communication through representatives
    constructed anew from image features at each layer, without predefined
    spatial groups or a separate latent sequence maintained across layers.

    \item We introduce RPAttention, an efficient global attention module
    built on learned representatives that can replace attention layers
    across standard vision backbones without changing the surrounding architecture.

    \item We demonstrate strong performance across four backbone
    families and three vision benchmarks, and characterize the behavior
    and practical efficiency of the representative bottleneck.
\end{itemize}

\section{Related Work}

\noindent\textbf{Efficient Vision Attention.}
Efficient attention methods reduce the cost of dense token interactions
through restricted connectivity, feature mappings, or compact
approximations.
Swin, CSWin, and NAT use windows or
neighborhoods~\citep{swin,cswin,nat}, while DAT and BiFormer adapt
sampling locations or select relevant regions from the
input~\citep{dat,biformer}.
Random Wins All~\citep{random_wins_all} instead uses fixed random
token groups, showing that efficient communication need not follow
a carefully designed spatial partition.
Linear attention methods reorganize computation through feature
mappings and shared statistics~\citep{linear_attention,performer},
with FLatten, PolaFormer, and InLine developing this direction for
vision~\citep{flatten_transformer,polaformer,inline_attention}.
Linformer and Nystr\"omformer use low-rank projections or landmark
approximations~\citep{linformer,nystromformer}.
RPAttention focuses on constructing explicit representatives from
input features and using them to mediate global communication.

\noindent\textbf{Spatial Proxy Construction.}
Several efficient vision transformers construct compact intermediate
tokens using spatial structure.
PVT and PVTv2 reduce the spatial resolution of keys and
values~\citep{pvt,pvt_v2}.
LRFormer~\citep{wu2025lrformer} applies self-attention to
fixed-size spatially pooled features and interpolates the result back to
the input resolution, while using depth-wise convolutions to capture local details.
Agent Attention~\citep{agent_attention} constructs agent tokens by
pooling query features, then uses them to aggregate global information
and broadcast it back to spatial queries.
Pooling may mix dissimilar features within one region,
while spatially distant features outside it do not contribute directly
to that region's agent query.
STViT~\citep{stvit} initializes super tokens on a spatial grid and
associates input tokens with nearby super tokens.
These methods introduce spatial priors through proxy construction
or local token association.
RPAttention instead computes competitive assignments from token
features to learned representative anchors and aggregates over
all spatial locations.
After representative interaction, spatial queries
retrieve the updated content through attention.
Its distinction lies in how representatives are constructed,
rather than global information access alone.

\noindent\textbf{Learned Aggregation and Latent Attention.}
Learning compact representations from visual features has been
explored in several forms.
TokenLearner~\citep{tokenlearner} generates adaptive tokens through
learned spatial weighting.
Slot Attention~\citep{slot_attention} uses competition across slots
and normalized aggregation, followed by iterative slot refinement
for object-centric representation learning.
GroupViT~\citep{groupvit} and kMaX-DeepLab~\citep{kmax_deeplab}
also learn grouping mechanisms for visual understanding and segmentation.
RPAttention shares the principle of feature-dependent aggregation,
but computes its assignments through a learned projection in a
single gathering step per block.
Its representatives support global communication rather than
being required to correspond to individual objects or classes.
Perceiver IO~\citep{perceiver_io} encodes inputs into a latent array
and decodes outputs through query-based cross-attention,
while BiXT~\citep{bixt} updates tokens and latents bidirectionally
using a shared similarity matrix.
RPAttention combines representatives constructed from the current
input, explicit interaction among them, and a separately computed
distribution map within an attention block that preserves
the spatial token interface.
\section{Methodology}
\label{sec:method}

\subsection{Overview}

Representative Attention (RPAttention) is an architectural replacement for attention modules in vision transformer backbones.
It preserves the standard input-output interface and the surrounding spatial backbone, 
but replaces direct token-to-token interaction with a compact, content-adaptive token-space bottleneck.

\begin{figure}[t]
  \centering
  \includegraphics[width=\linewidth]{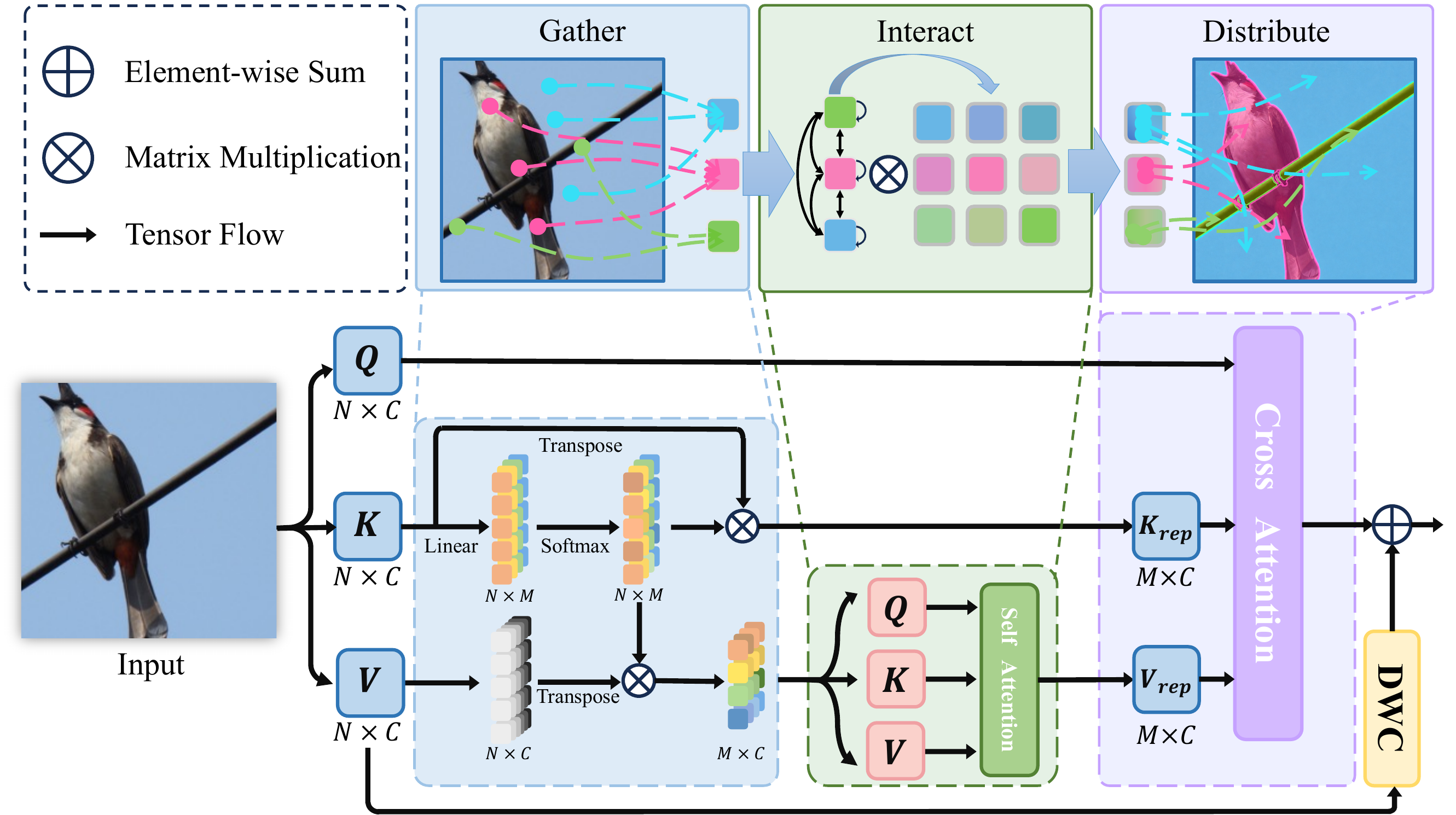}
  \caption{Overview of Representative Attention.
Spatial tokens are gathered into a compact set of representatives according to representation similarity (\textit{Gather}).
These tokens then model global dependencies through lightweight self-attention in the compact token space (\textit{Interact}).
Finally, the refined representations are distributed back to all spatial tokens through query-driven cross-attention (\textit{Distribute}).
A parallel depth-wise convolution branch preserves local structural details.}
  \label{fig:architecture}
\end{figure}

Given an input feature map
$\mathbf{X} \in \mathbb{R}^{B \times N \times C}$, where $B$ is the batch
size, $N$ is the number of spatial tokens, and $C$ is the channel dimension,
RPAttention first computes the standard query, key, and value projections and
reshapes them into $h$ heads:
\begin{equation}
    \mathbf{Q} = \mathbf{X}\mathbf{W}_q,\qquad
    \mathbf{K} = \mathbf{X}\mathbf{W}_k,\qquad
    \mathbf{V} = \mathbf{X}\mathbf{W}_v,\quad
    \mathbf{Q},\mathbf{K},\mathbf{V}
    \in \mathbb{R}^{B \times h \times N \times d},
\end{equation}

where $C=hd$. For reference, standard global self-attention computes
\begin{equation}
    \operatorname{Attention}^{S}(\mathbf{Q},\mathbf{K},\mathbf{V})
    =
    \operatorname{Softmax}\left(
    \frac{\mathbf{Q}\mathbf{K}^{\top}}{\sqrt{d}}
    \right)\mathbf{V},
\end{equation}
which explicitly forms pairwise interactions among all $N$ spatial tokens.
RPAttention instead factors global communication through $M \ll N$ representatives:
\begin{equation}
\mathbf{O}^{\mathrm{RP}}
=
\underbrace{
\operatorname{Distribute}\left(
\mathbf{Q},
\RPKL,
\operatorname{Interact}(\RPVL)
\right)
}_{\text{spatial queries retrieve representative context}},
\qquad
\{\RPKL,\RPVL\}
=
\underbrace{
\operatorname{Gather}(\mathbf{K},\mathbf{V})
}_{\text{tokens form representatives}}.
\end{equation}
This formulation preserves the standard attention interface while constructing intermediate tokens through representation similarity rather than coordinate-based reduction. Figure~\ref{fig:architecture} provides an overview of the complete Representative Attention module.

\subsection{Representation-Conditioned Gathering}

The gather step determines how spatial tokens are coarsened into a compact
token-space bottleneck. Unlike fixed grid pooling or strided reduction,
RPAttention constructs the representatives from learned feature
compatibilities within each attention head. Let
$\mathbf{W}_g \in \mathbb{R}^{d \times M}$ denote the gather projection, where
$M \ll N$ is the number of representatives.
The gather logits are computed from the key features as
\begin{equation}
    \mathbf{A}
    =
    \operatorname{Softmax}_{M}\!\left(\mathbf{K}\mathbf{W}_g\right),
    \qquad
    \mathbf{A}\in \mathbb R^{B\times h\times N\times M},
\end{equation}
where the softmax is applied over the representative dimension $M$. For each
attention head, the row $\mathbf{A}_{n,:}$ defines a distribution over the
representatives for the $n$-th spatial token. Thus, $A_{n,m}$ specifies the
fractional contribution of token $n$ to representative $m$. It is determined
by the learned feature compatibility between the token key and the
corresponding representative anchor, namely the $m$-th column of
$\mathbf{W}_g$. Consequently,
\begin{equation}
    A_{n,m}\geq 0,
    \qquad
    \sum_{m=1}^{M}A_{n,m}=1.
    \label{eq:gather_assignment}
\end{equation}
Each token can therefore contribute to multiple representatives, while the
competition across the $M$ representatives determines how its contribution is
distributed. Because this routing depends on feature compatibility rather
than token coordinates or predefined spatial regions, tokens at arbitrary
spatial locations can contribute to the same representative.
Different representatives may receive different total assignment masses.
We normalize the incoming weights of each representative to keep
the aggregated feature scales comparable:
\begin{equation}
    \hat{\mathbf{A}}_{n,m}
    =
    \frac{\mathbf{A}_{n,m}}
    {\sum_{t=1}^{N}\mathbf{A}_{t,m}+\epsilon}.
\end{equation}
In contrast to the row-normalized assignment matrix $\mathbf{A}$, the
normalization of $\hat{\mathbf{A}}$ is performed over tokens for each
representative. It converts the incoming assignment weights into
representative-wise aggregation coefficients.
This column-wise normalization makes the representative features less
sensitive to the total assignment mass of each representative and keeps their
scales comparable across the representative set.
The normalized assignments are then used to construct the representative
keys and values:
\begin{equation}
    \RPKL
    =
    \hat{\mathbf{A}}^{\top}\mathbf{K},
    \qquad
    \RPVL
    =
    \hat{\mathbf{A}}^{\top}\mathbf{V},
    \qquad
    \RPKL,\RPVL
    \in\mathbb{R}^{B\times h\times M\times d}.
\end{equation}
The representative keys $\RPKL$ define the routing geometry used to address
the compact latent space, whereas the representative values $\RPVL$ carry the
aggregated visual content that is later refined and distributed back to the
spatial queries. This feed-forward construction produces input-derived representatives in a
single pass. The representatives are therefore not fixed latent vectors:
their values depend on the current token features and on the learned
compatibility projection.

\begin{figure}[t]
  \centering
  \includegraphics[width=\linewidth]{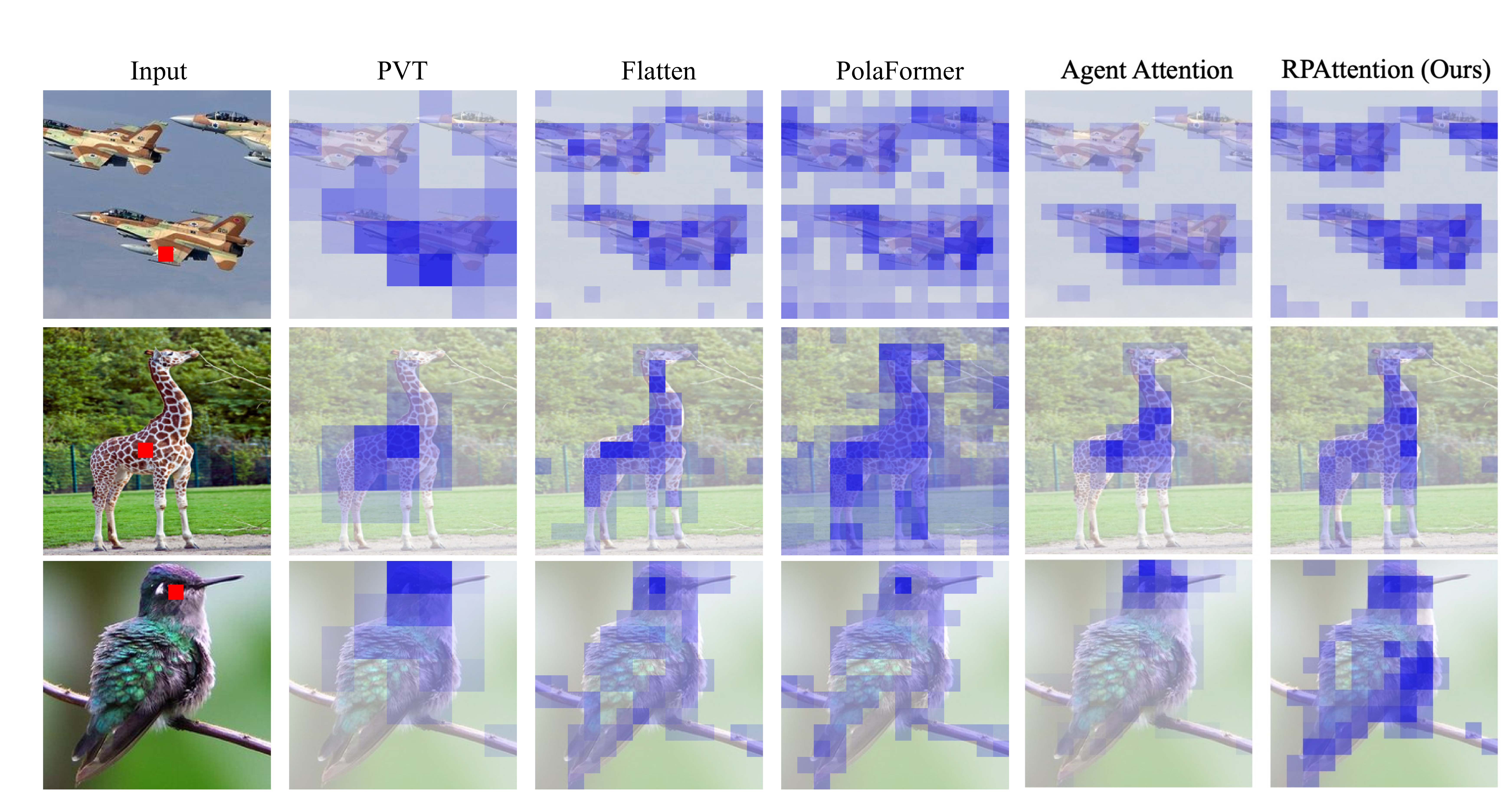}
  \caption{Qualitative comparison of query-to-space routing.
    Red squares mark query locations, and blue overlays show the corresponding
    response maps.}
  \label{fig:analysis}
\end{figure}

\subsection{Latent Interaction}

After gathering, RPAttention performs global interaction among the $M$
representatives rather than among all $N$ spatial tokens. We apply Layer Normalization to the representative keys and values,
yielding $\RPbarKL=\operatorname{LN}_{K}(\RPKL)$ and
$\RPbarVL=\operatorname{LN}_{V}(\RPVL)$.
This normalizes the aggregated features before they are
used for query-to-representative retrieval and representative interaction. The latent
interaction projections are derived from the representative values:
\begin{equation}
    \widetilde{\mathbf{Q}}_L
    =
    \RPbarVL \mathbf{W}^{L}_{q},
    \qquad
    \widetilde{\mathbf{K}}_L
    =
    \RPbarVL \mathbf{W}^{L}_{k},
    \qquad
    \widetilde{\mathbf{V}}_L
    =
    \RPbarVL \mathbf{W}^{L}_{v}.
\end{equation}
The refined representative content is then obtained by self-attention within
the compact latent space:
\begin{equation}
    \mathbf{Z}_L
    =
    \RPbarVL
    +
    \operatorname{Softmax}\left(
    \frac{
    \widetilde{\mathbf{Q}}_L
    \widetilde{\mathbf{K}}_L^{\top}
    }{\sqrt{d}}
    \right)
    \widetilde{\mathbf{V}}_L.
\end{equation}
The interaction projections are derived from $\RPbarVL$, leaving $\RPbarKL$ unchanged for query-conditioned distribution.
This design lets representative values exchange content
while retaining the gathered keys for spatial queries to retrieve the updated
representations.
Because the interaction is performed among only $M$ representatives, this
stage has a quadratic cost in $M$ rather than in the number of spatial tokens
$N$.
\subsection{Query-Conditioned Distribution}
The refined representative content is returned to the spatial tokens
through query-conditioned attention. Each spatial query attends to the
normalized representative keys $\RPbarKL$:
\begin{equation}
    \mathbf{O}_{\mathrm{global}}
    =
    \operatorname{Softmax}\left(
    \frac{\mathbf{Q}\RPbarKL^{\top}}{\sqrt{d}}
    \right)\mathbf{Z}_L.
\end{equation}
The distribution map is computed separately from, rather than taken directly
from, the gathering assignments. Thus, a token's contribution to a
representative need not determine how it retrieves the refined content.
The compact global bottleneck may weaken local boundaries and other
fine-grained details. To preserve this information, we introduce a parallel
local bypass implemented with a depth-wise convolution (DWC), following
FLatten Transformer~\citep{flatten_transformer}. The global and local paths
are fused before the output projection:
\begin{equation}
\begin{aligned}
   \mathbf{X}_{\mathrm{out}}
    =
    \left(
    \operatorname{MergeHeads}(\mathbf{O}_{\mathrm{global}})
    +
    \operatorname{DWC}(\mathbf{X}\mathbf{W}_v)
    \right)\mathbf{W}_o.
\end{aligned}
\end{equation}
Here, $\operatorname{MergeHeads}$ concatenates the $h$ attention heads into $C$ channels.
The representative path captures global context, while the DWC path preserves
local spatial details. Their fusion combines long-range interaction with local
structure preservation without changing the spatial input--output interface.

\subsection{Complexity Analysis}
RPAttention can replace an attention module without changing the surrounding
transformer layer, patch embedding, MLP, or residual structure.
Let $N$ denote the number of spatial tokens, $C=hd$ the channel dimension,
and $M$ the number of representatives.
Ignoring lower-order normalization and dropout terms, the computational cost
of one RPAttention block is
\begin{equation}
\begin{aligned}
\Omega_{\mathrm{RP}}
={}&
\underbrace{4NC^2}_{\text{Projections}}
+
\underbrace{3NMC}_{\text{Gather}}
+
\underbrace{3MCd + 2M^2C}_{\text{Interaction}}
+
\underbrace{2NMC}_{\text{Distribute}}
+
\underbrace{k^2NC}_{\text{DWC}},
\end{aligned}
\end{equation}
where $C=hd$ is the channel dimension and $k=3$ is the kernel size of the
depth-wise convolution.

The first term accounts for the three input projections and the output
projection. In the gathering stage, the projection
$\mathbf{K}\mathbf{W}_g$ costs $NMC$, while aggregating the keys and values
into representatives costs an additional $2NMC$. The latent interaction requires $3MCd$ for the head-wise latent QKV
projections and $2M^2C$ for the representative-level attention matrix
and value aggregation. 
The distribution stage contributes another $2NMC$ for computing query-to-
representative affinities and reading out the refined representative
content.
For comparison, standard global self-attention has cost
\begin{equation}
    \Omega_{\mathrm{SA}}
    =
    4NC^2 + 2N^2C,
\end{equation}
where the second term accounts for the $N\times N$ attention affinities and
the corresponding value aggregation.
Thus, RPAttention replaces the quadratic token-interaction term $2N^2C$
with
\begin{equation}
    5NMC + 3MCd + 2M^2C + k^2NC.
    \label{eq:linear_interaction_cost}
\end{equation}
When $M$, $C$, and $k$ are fixed with respect to $N$, the token-dependent
complexity of RPAttention scales linearly with the number of spatial tokens:
\begin{equation}
    \Omega_{\mathrm{RP}}
    =
    \mathcal{O}\!\left(
    NC^2 + NMC + MCd + M^2C + k^2NC
    \right)
    =
    \mathcal{O}(N).
\end{equation}
Unlike local or windowed attention, this linear scaling does not require
restricting queries to spatial neighborhoods. Every spatial query can
access the same compact set of globally gathered representatives,
preserving a global communication path.

\section{Experimental Validation}
\label{sec:experiments}
\subsection{Experimental Setup}
We organize our experiments into three categories: image classification,
dense prediction, and practical efficiency. ImageNet-1K
~\citep{imagenet} is used for classification, COCO~\citep{coco} for object detection and instance segmentation, and ADE20K
~\citep{ade20k} for semantic segmentation. We further
measure inference latency and throughput at different input resolutions to
evaluate the practical scaling behavior of RPAttention.

We instantiate RPAttention as a
replacement for the attention blocks in four backbone families:
DeiT~\citep{deit}, PVT~\citep{pvt}, PVTv2~\citep{pvt_v2}, and Swin
Transformer~\citep{swin}. All other backbone components and task-specific prediction heads
are kept unchanged. The main tables compare each modified backbone
with its original attention-based counterpart.
Detailed training settings are provided in
Appendix~\ref{app:training}, and measurement protocols and additional
practical-efficiency comparisons in
Appendix~\ref{sec:efficiency_protocol}.

\subsection{ImageNet-1K Classification}

\begin{figure}[t]
    \centering
    \includegraphics[width=\linewidth]
    {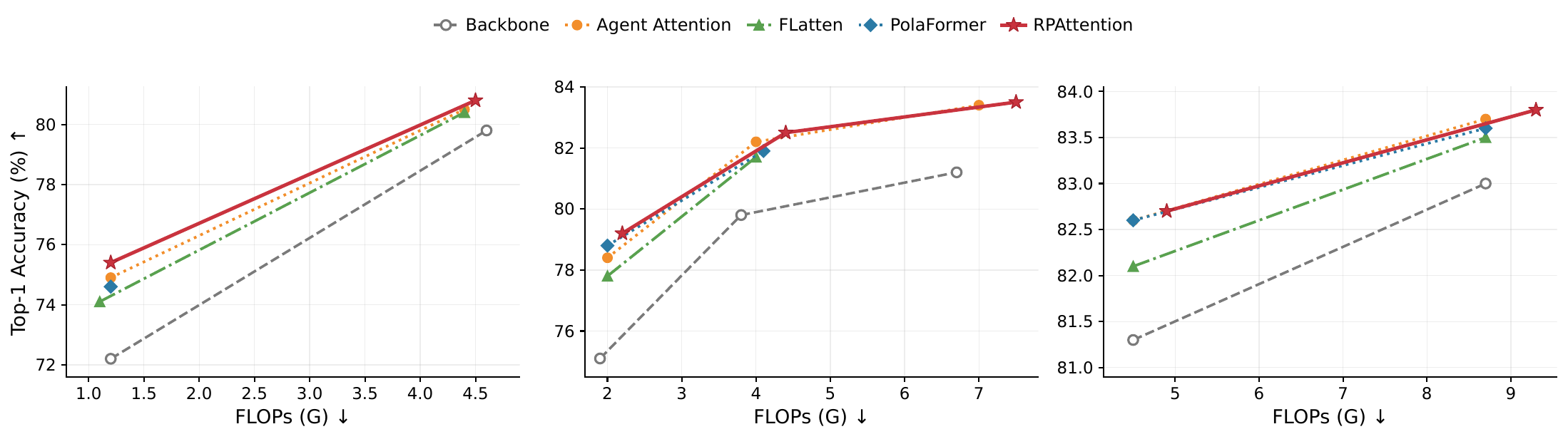}
    \caption{ImageNet-1K accuracy--FLOPs comparison with efficient-attention methods. From left to right: DeiT, PVT, and Swin. 
    Points along each curve follow increasing model size.}
    \label{fig:imagenet_tradeoff}
\end{figure}

Table~\ref{tab:imagenet_main} reports ImageNet-1K classification results
across four backbone families and ten model configurations. 
RPAttention consistently improves Top-1 accuracy over the corresponding
backbone across all evaluated configurations, with gains ranging from
$+0.6$ to $+4.1$ percentage points. For example,
RPAttention improves PVT-T/PVT-S/PVT-M from $75.1/79.8/81.2$ to
$79.2/82.5/83.5$, respectively, and improves Swin-S from $83.0$ to $83.8$.
These results demonstrate that the proposed method is effective across both flat and hierarchical vision transformer
backbones. 
Figure~\ref{fig:imagenet_tradeoff} further compares RPAttention with
other efficient-attention methods on DeiT, PVT, and Swin.
Detailed numerical comparisons are provided in
Appendix~\ref{app:classification_comparisons}.

Figure~\ref{fig:analysis} provides a qualitative comparison of
query-to-space routing.
RPAttention responds to spatially separated aircraft regions and extends
the bird-head query to the body, whereas PVT and Agent Attention show
more localized responses in the bird example.
In the aircraft and giraffe examples, RPAttention also exhibits less
background spread than PolaFormer.
These examples illustrate how the representative bottleneck can support
broad, content-dependent responses while retaining object-related
spatial structure.
Additional visualizations are provided in Appendix~\ref{app:qualitative}.
\begin{table}[t]
\centering
\caption{ImageNet-1K classification results across four backbone families.}
\label{tab:imagenet_main}

\begin{minipage}[t]{0.49\linewidth}
\centering
\begingroup
\setlength{\tabcolsep}{0.8mm}
\setlength{\aboverulesep}{0pt}
\setlength{\belowrulesep}{0pt}
\renewcommand{\arraystretch}{1.08}

\begin{tabular}[t]{@{}lccc@{}}
\toprule
\textbf{Method} &
\textbf{Params} & \textbf{FLOPs} & \textbf{Top-1} \\
\midrule
DeiT-T & 5.7M & 1.2G & 72.2 \\
\rowcolor{verylightgray}{ + RPAttention}
& {5.9M} & {1.2G} & {\textbf{75.4 (+3.2)}} \\
\midrule
DeiT-S & 22.1M & 4.6G & 79.8 \\
\rowcolor{verylightgray}{ + RPAttention}
& {22.3M} & {4.5G} & {\textbf{80.8 (+1.0)}} \\
\midrule
PVT-T & 13.2M & 1.9G & 75.1 \\
\rowcolor{verylightgray}{ + RPAttention}
& {11.5M} & {2.2G} & {\textbf{79.2 (+4.1)}} \\
\midrule
PVT-S & 24.5M & 3.8G & 79.8 \\
\rowcolor{verylightgray}{ + RPAttention}
& {20.5M} & {4.4G} & {\textbf{82.5 (+2.7)}} \\
\midrule
PVT-M & 44.2M & 6.7G & 81.2 \\
\rowcolor{verylightgray}{ + RPAttention}
& {35.5M} & {7.5G} & {\textbf{83.5 (+2.3)}} \\
\bottomrule
\end{tabular}
\endgroup
\end{minipage}%
\hfill
\begin{minipage}[t]{0.49\linewidth}%
\centering
\begingroup
\setlength{\tabcolsep}{0.8mm}
\setlength{\aboverulesep}{0pt}
\setlength{\belowrulesep}{0pt}
\renewcommand{\arraystretch}{1.08}

\begin{tabular}[t]{@{}lccc@{}}
\toprule
\textbf{Method} &
\textbf{Params} & \textbf{FLOPs} & \textbf{Top-1} \\
\midrule
PVTv2-B0 & 3.7M & 0.5G & 70.5 \\
\rowcolor{verylightgray}{ + RPAttention}
& {3.2M} & {0.7G} & {\textbf{73.1 (+2.6)}} \\
\midrule
PVTv2-B1 & 13.1M & 2.1G & 78.7 \\
\rowcolor{verylightgray}{ + RPAttention}
& {12.3M} & {2.4G} & {\textbf{80.5 (+1.8)}} \\
\midrule
PVTv2-B2 & 25.4M & 4.0G & 82.0 \\
\rowcolor{verylightgray}{ + RPAttention}
& {21.3M} & {4.6G} & {\textbf{82.6 (+0.6)}} \\
\midrule
Swin-T & 29.0M & 4.5G & 81.3 \\
\rowcolor{verylightgray}{ + RPAttention}
& {28.4M} & {4.9G} & {\textbf{82.7 (+1.4)}} \\
\midrule
Swin-S & 50M & 8.7G & 83.0 \\
\rowcolor{verylightgray}{ + RPAttention}
& {50M} & {9.3G} & {\textbf{83.8 (+0.8)}} \\
\bottomrule
\end{tabular}
\endgroup
\end{minipage}

\end{table}

\begin{table}[t]
\centering
\caption{Dense prediction results on COCO and ADE20K.
FLOPs are computed at $1280\times800$ for COCO and $512\times2048$
for ADE20K. Unavailable entries are denoted by ``--''.}
\label{tab:dense_prediction}

\begingroup
\setlength{\tabcolsep}{0.8mm}
\renewcommand{\arraystretch}{1.08}

\resizebox{\linewidth}{!}{%
\begin{tabular}{@{}lccccccccccc@{}}
\toprule
\multirow{2}{*}{\textbf{Method}}
& \multicolumn{7}{c}{\textbf{Detection \& Instance Seg.}}
& \multicolumn{4}{c}{\textbf{Semantic Seg.}} \\
\cmidrule(lr){2-8}
\cmidrule(l){9-12}
& \textbf{FLOPs}
& \textbf{AP$^b$}
& \textbf{AP$^b_{50}$}
& \textbf{AP$^b_{75}$}
& \textbf{AP$^m$}
& \textbf{AP$^m_{50}$}
& \textbf{AP$^m_{75}$}
& \textbf{FLOPs}
& \textbf{Params}
& \textbf{mIoU}
& \textbf{mAcc} \\
\midrule

PVT-T
& 240G & 36.7 & 59.2 & 39.3 & 35.1 & 56.7 & 37.3
& 158G & 17M & 36.57 & 46.72 \\
\rowcolor{verylightgray}
{ + RPAttention}
& 215G & 40.4 & 62.8 & 43.8 & 37.7 & 59.9 & 40.4
& 128G & 15M & 40.00 & 53.54 \\
\midrule

PVT-S
& 305G & 40.4 & 62.9 & 43.8 & 37.8 & 60.1 & 40.3
& 225G & 28M & 41.95 & 53.02 \\
\rowcolor{verylightgray}
{ + RPAttention}
& 259G & 44.1 & 66.6 & 48.5 & 40.8 & 63.7 & 44.3
& 177G & 24M & 43.25 & 56.04 \\
\midrule

PVT-M
& 392G & 42.0 & 64.4 & 45.6 & 39.0 & 61.6 & 42.1
& 315G & 48M & 42.91 & 53.80 \\
\rowcolor{verylightgray}
{ + RPAttention}
& 322G & 45.9 & 68.2 & 50.4 & 41.9 & 65.4 & 45.3
& 241G & 39M & 45.16 & 58.13 \\
\midrule

Swin-T
& 267G & 43.7 & 66.6 & 47.7 & 39.8 & 63.3 & 42.7
& 945G & 60M & 44.51 & 55.61 \\
\rowcolor{verylightgray}
{ + RPAttention}
& 271G & 44.5 & 67.3 & 48.4 & 40.5 & 64.2 & 43.3
& 949G & 60M & 46.97 & 59.17 \\
\midrule

Swin-S
& 358G & 45.7 & 67.9 & 50.4 & 41.1 & 64.9 & 44.2
& 1038G & 81M & 47.64 & 58.78 \\
\rowcolor{verylightgray}
{ + RPAttention}
& 360G & 46.3 & 69.0 & 50.4 & 41.8 & 65.7 & 45.1
& 1042G & 81M & 48.48 & 60.12 \\
\midrule

PVTv2-B0
& 196G & 38.2 & 60.5 & 40.7 & 36.2 & 57.8 & 38.6
& 113G & 7.6M & 37.2 & -- \\
\rowcolor{verylightgray}
{ + RPAttention}
& 184G & 38.7 & 61.0 & 41.4 & 36.1 & 57.9 & 38.5
& 100G & 7.1M & 37.93 & 49.08 \\
\midrule

PVTv2-B1
& 244G & 41.8 & 64.3 & 45.9 & 38.8 & 61.2 & 41.6
& 162G & 17.8M & 42.5 & -- \\
\rowcolor{verylightgray}
{ + RPAttention}
& 220G & 43.2 & 65.6 & 47.3 & 39.8 & 62.5 & 42.7
& 136G & 16.0M & 43.10 & 54.86 \\
\midrule

PVTv2-B2
& 309G & 45.3 & 67.1 & 49.6 & 41.2 & 64.2 & 44.4
& 230G & 29.1M & 45.2 & -- \\
\rowcolor{verylightgray}
{ + RPAttention}
& 264G & 46.2 & 68.1 & 50.6 & 42.0 & 64.9 & 45.3
& 182G & 25.1M & 46.69 & 58.54 \\

\bottomrule
\end{tabular}%
}

\endgroup
\end{table}

\subsection{Dense Prediction}

We evaluate RPAttention on COCO 2017~\citep{coco} for object detection
and instance segmentation using Mask R-CNN~\citep{mask_rcnn} with the
$1\times$ schedule, and on ADE20K~\citep{ade20k} for semantic segmentation.
Table~\ref{tab:dense_prediction} compares RPAttention with the corresponding
backbone baselines across the evaluated configurations.
Additional comparisons on COCO and ADE20K are provided
in Appendix~\ref{app:dense_comparisons}.

\noindent\textbf{Object detection and instance segmentation.}
RPAttention improves box AP across all evaluated configurations by
$0.5$--$3.9$ points and generally improves mask AP.
The largest gains are $+3.9$ box AP on PVT-M and $+3.0$ mask AP on PVT-S,
while PVTv2-B0 shows a minor $0.1$ mask AP decrease.
PVT and PVTv2 variants also use fewer FLOPs than their corresponding backbones.

\noindent\textbf{Semantic segmentation.}
RPAttention improves mIoU across all eight configurations by
$0.60$--$3.43$ points.
For PVT and PVTv2, these gains are accompanied by fewer parameters and lower
FLOPs, while Swin improves mIoU at a similar parameter count and computational cost.

\begin{figure}[t]
    \centering
    \includegraphics[width=0.80\linewidth]
    {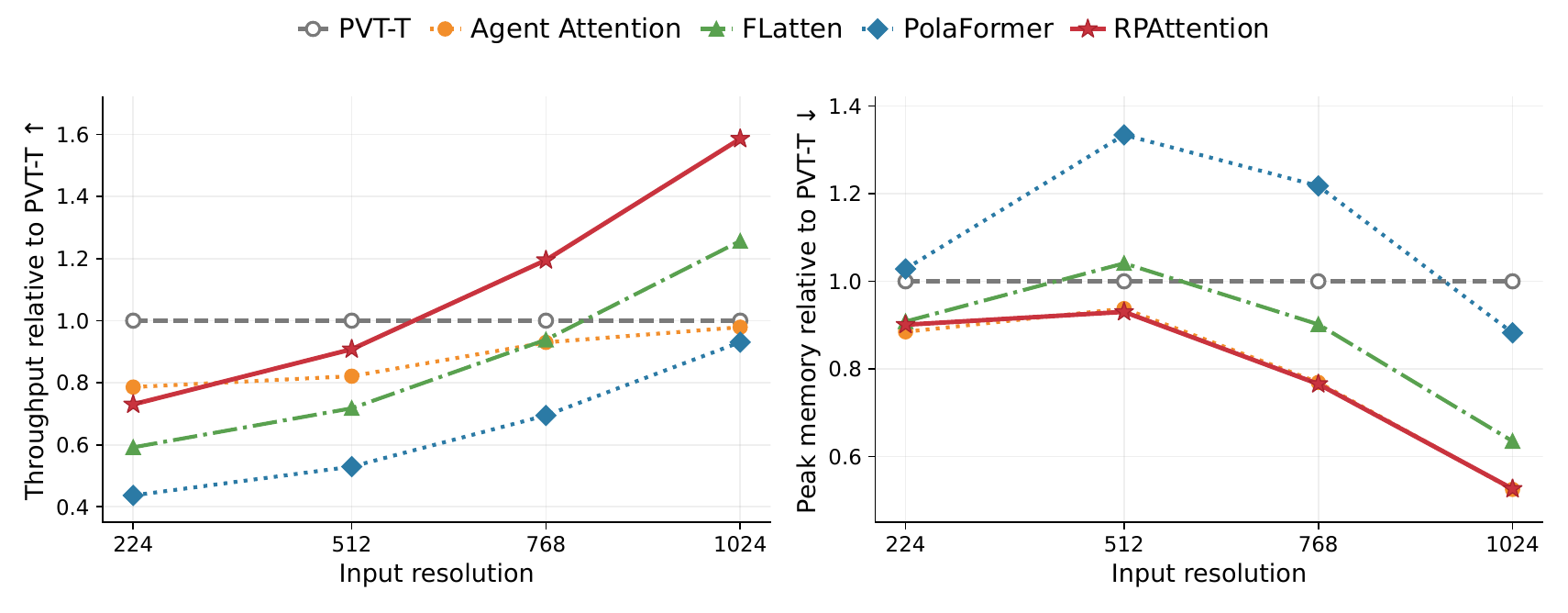}

    \vspace{0.6em}

    \includegraphics[width=0.80\linewidth]
    {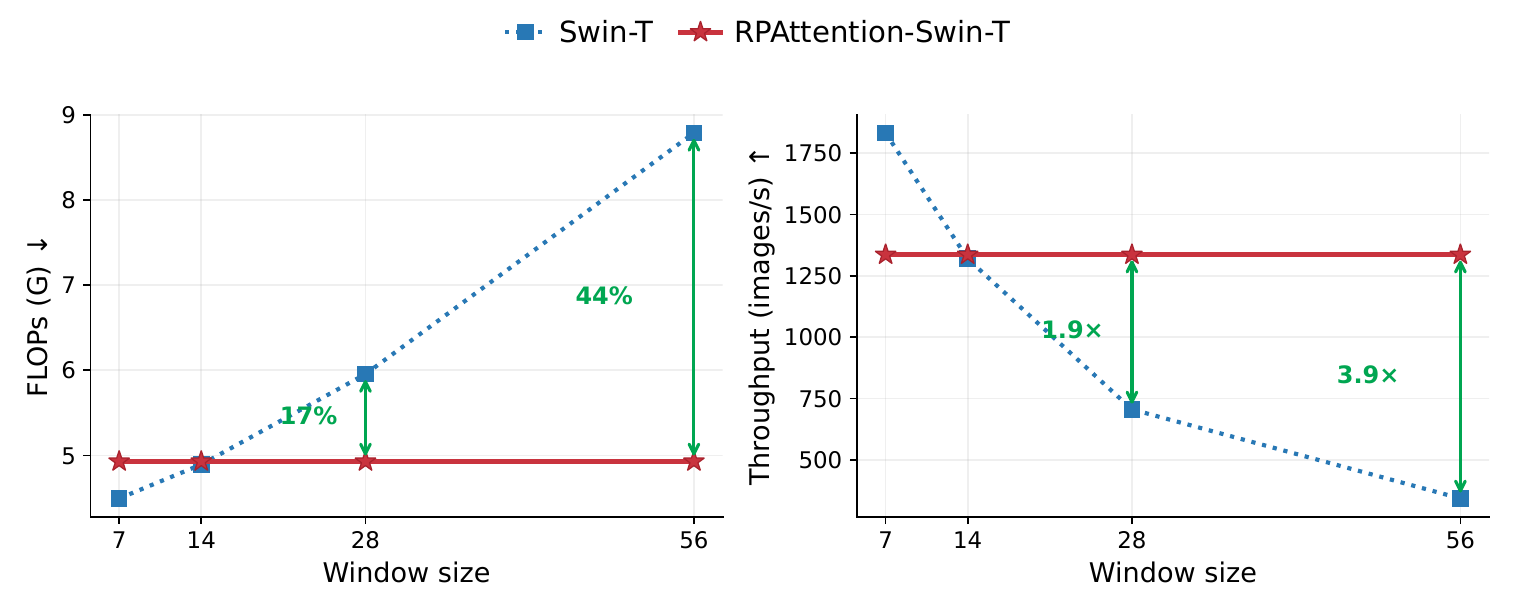}

    \caption{Practical efficiency on an RTX 3090.
    Top: matched-batch throughput and batch-1 peak GPU memory
    across input resolutions, normalized to PVT-T.
    Bottom: FLOPs and throughput as the Swin-T attention window
    grows, with RPAttention-Swin-T held fixed.}
    \label{fig:efficiency_scaling}
\end{figure}

\noindent\textbf{Efficiency scaling.}
We further examine practical efficiency under increasing spatial workloads.
As input resolution grows, RPAttention-PVT-T becomes faster than
PVT-T and the compared efficient-attention variants, while using
less peak memory than PVT-T
(Figure~\ref{fig:efficiency_scaling}, top).
Its peak memory is comparable to Agent Attention.
A complementary study varies the attention-window size of Swin-T
while keeping RPAttention-Swin-T fixed, showing increasingly favorable
FLOPs and throughput comparisons as the window size grows
(Figure~\ref{fig:efficiency_scaling}, bottom).

\subsection{Ablation Studies}
\begin{table}[t]
\centering
\begin{minipage}[t]{0.53\linewidth}
\vspace{0pt}
\centering
\caption{Component ablations on PVTv2-B0.}
\label{tab:ablation_components}
\normalsize
\setlength{\tabcolsep}{3pt}
\begin{tabular}{@{}lccc@{}}
\toprule
\textbf{Variant} & \textbf{Params} & \textbf{FLOPs} & \textbf{Top-1} \\
\midrule
PVTv2-B0          & 3.7M  & 0.5G  & 70.5 \\
w/o DWC           & 3.24M & 0.71G & 72.7 (+2.2) \\
w/ k-means routing & 3.29M & 0.78G & 72.6 (+2.1) \\
w/o Interaction   & 3.23M & 0.70G & 72.8 (+2.3) \\
\rowcolor{verylightgray}
RPAttention       & 3.25M & 0.72G & \textbf{73.1 (+2.6)} \\
\bottomrule
\end{tabular}
\end{minipage}\hfill
\begin{minipage}[t]{0.45\linewidth}
\vspace{0pt}
\centering
\caption{Representative counts on PVTv2-B0.}
\label{tab:representative_counts}
\normalsize
\setlength{\tabcolsep}{3pt}
\begin{tabular}{@{}lccc@{}}
\toprule
\textbf{Reps. per stage} & \textbf{Top-1}
& \textbf{mIoU} & \textbf{mAcc} \\
\midrule
$(9,16,25,25)$  & 72.8 & -- & -- \\
$(9,16,49,49)$  & \textbf{73.1} & 37.21 & 48.72 \\
$(49,49,49,49)$ & 72.9 & -- & -- \\
$(96,64,64,49)$ & 73.0 & -- & -- \\
\rowcolor{verylightgray}
$(96,64,49,49)$ & \textbf{73.1}
& \textbf{37.93} & \textbf{49.08} \\
\bottomrule
\end{tabular}
\end{minipage}
\end{table}
\noindent\textbf{Component analysis.}
Table~\ref{tab:ablation_components} examines the contributions of
local refinement, representative interaction, and learned gathering
on PVTv2-B0, with each variant compared against the full model.
Removing the DWC branch reduces Top-1 accuracy from 73.1\% to 72.7\%,
while retaining a 2.2-point gain over the original backbone.
Thus, the improvement does not rely solely on the local convolutional
branch, although local refinement provides an additional benefit.
Removing representative interaction reduces accuracy to 72.8\%,
suggesting that exchanging information among representatives is
beneficial beyond gathering and distribution alone.
We replace the learned projection in Gather with per-input
k-means routing using three iterations, leaving the remaining
components unchanged.
This variant achieves 72.6\% accuracy at 0.78G FLOPs,
whereas learned gathering achieves 73.1\% at 0.72G,
providing higher accuracy at lower computational cost.

\noindent\textbf{Number of representatives.}
Table~\ref{tab:representative_counts} compares five stage-wise
representative configurations on PVTv2-B0.
ImageNet-1K Top-1 accuracy varies only from 72.8\% to 73.1\%.
Early stages operate at higher spatial resolutions, where a small
representative budget may overly compress spatial detail needed
for dense prediction.
With the last two stages fixed at $(49,49)$, increasing the
early-stage counts from $(9,16)$ to $(96,64)$ maintains 73.1\%
Top-1 accuracy while improving ADE20K mIoU by 0.72 points
and mAcc by 0.36 points.
These results motivate our default choice of $(96,64,49,49)$.

Additional analyses of cross-seed routing stability, foreground alignment,
and query-response visualizations are provided in
Appendices~\ref{app:representative_behavior},
\ref{app:foreground}, and~\ref{app:qualitative}, respectively.

\section{Conclusion}

We introduced Representative Attention (RPAttention), an efficient global
attention module that constructs compact representatives from input
features rather than fixed spatial partitions.
Its \textit{Gather--Interact--Distribute} pipeline gathers tokens through
learned feature compatibility, enables interaction among representatives,
and distributes the refined information according to each query,
while preserving the spatial token interface.
With $M$ representatives and $N$ spatial tokens, the cost of token
interactions is $\mathcal{O}(NM + M^2)$.
Experiments on ImageNet-1K classification, COCO object detection and
instance segmentation, and ADE20K semantic segmentation demonstrate
improved overall performance over the corresponding backbone baselines
and competitive results against recent efficient attention methods.
These findings support learning token compression from visual features
as an effective alternative to spatially defined proxy construction
for global visual communication.

\subsection*{Reproducibility Statement} 
Appendix~\ref{app:implementation} provides a PyTorch-style implementation
of the core Gather--Interact--Distribute module.
Appendix~\ref{app:model_architectures} lists the stage-wise architectures
and representative counts of all evaluated models.
Appendix~\ref{app:training} documents the training and evaluation recipes
for ImageNet-1K, COCO, and ADE20K, including optimization and augmentation
settings.
For efficiency measurements, Appendix~\ref{sec:efficiency_protocol}
details the runtime environment and measurement protocol.
Upon acceptance, we will release the complete training and evaluation
code, configurations, scripts, and principal checkpoints.

\subsection*{AI Use Statement}
AI tools were used to polish the language and improve readability.
The authors thoroughly reviewed the revised text and take full
responsibility for the final manuscript.

\bibliography{references}

@inproceedings{linear_attention,
  title={Transformers are {RNNs}: Fast Autoregressive Transformers with Linear Attention},
  author={Katharopoulos, Angelos and Vyas, Apoorv and Pappas, Nikolaos and Fleuret, Fran{\c{c}}ois},
  booktitle={ICML},
  year={2020}
}

@inproceedings{vit,
  title={An Image is Worth 16x16 Words: Transformers for Image Recognition at Scale},
  author={Dosovitskiy, Alexey and Beyer, Lucas and Kolesnikov, Alexander and Weissenborn, Dirk and Zhai, Xiaohua and Unterthiner, Thomas and Dehghani, Mostafa and Minderer, Matthias and Heigold, Georg and Gelly, Sylvain and others},
  booktitle={ICLR},
  year={2021}
}

@inproceedings{deit,
  title={Training data-efficient image transformers \& distillation through attention},
  author={Touvron, Hugo and Cord, Matthieu and Douze, Matthijs and Massa, Francisco and Sablayrolles, Alexandre and J{\'e}gou, Herv{\'e}},
  booktitle={ICML},
  year={2021}
}

@inproceedings{pvt,
  title={Pyramid vision transformer: A versatile backbone for dense prediction without convolutions},
  author={Wang, Wenhai and Xie, Enze and Li, Xiang and Fan, Deng-Ping and Song, Kaitao and Liang, Ding and Lu, Tong and Luo, Ping and Shao, Ling},
  booktitle={ICCV},
  year={2021}
}

@article{pvt_v2,
  title={{PVT v2}: Improved baselines with pyramid vision transformer},
  author={Wang, Wenhai and Xie, Enze and Li, Xiang and Fan, Deng-Ping and Song, Kaitao and Liang, Ding and Lu, Tong and Luo, Ping and Shao, Ling},
  journal={Computational Visual Media},
  year={2022}
}

@inproceedings{swin,
  title={{Swin} transformer: Hierarchical vision transformer using shifted windows},
  author={Liu, Ze and Lin, Yutong and Cao, Yue and Hu, Han and Wei, Yixuan and Zhang, Zheng and Lin, Stephen and Guo, Baining},
  booktitle={ICCV},
  year={2021}
}

@inproceedings{cswin,
  title={{CSWin} transformer: A general vision transformer backbone with cross-shaped windows},
  author={Dong, Xiaoyi and Bao, Jianmin and Chen, Dongdong and Zhang, Weiming and Yu, Nenghai and Yuan, Lu and Chen, Dong and Guo, Baining},
  booktitle={CVPR},
  year={2022}
}

@inproceedings{nat,
  title={Neighborhood Attention Transformer},
  author={Hassani, Ali and Walton, Steven and Li, Jiachen and Li, Shen and Shi, Humphrey},
  booktitle={CVPR},
  year={2023}
}

@inproceedings{
    bixt,
    title={Perceiving Longer Sequences With Bi-Directional Cross-Attention Transformers},
    author={Markus Hiller and Krista A. Ehinger and Tom Drummond},
    booktitle={NeurIPS},
    year={2024},
}

@inproceedings{tokenlearner,
  title={TokenLearner: Adaptive Space-Time Tokenization for Videos},
  author={ Ryoo, Michael and Piergiovanni, AJ and Arnab, Anurag and Dehghani, Mostafa and Angelova, Anelia},
  booktitle={NeurIPS},
  year={2021}
}

@inproceedings{random_wins_all,
  title={Random Wins All: Rethinking Grouping Strategies for Vision Tokens},
  author={Fan, Qihang and Ai, Yuang and Huang, Huaibo and He, Ran},
  booktitle={CVPR},
  year={2026}
}

@InProceedings{mae,
    author    = {He, Kaiming and Chen, Xinlei and Xie, Saining and Li, Yanghao and Doll\'ar, Piotr and Girshick, Ross},
    title     = {Masked Autoencoders Are Scalable Vision Learners},
    booktitle = {CVPR},
    year      = {2022}
}

@inproceedings{stvit,
  title={Vision Transformer with Super Token Sampling},
  author={Huang, Huaibo and Zhou, Xiaoqiang and Cao, Jie and He, Ran and
          Tan, Tieniu},
  booktitle={CVPR},
  year={2023}
}

@inproceedings{performer,
  title={Rethinking Attention with Performers},
  author={Choromanski, Krzysztof and Likhosherstov, Valerii and Dohan, David and Song, Xingyou and Gane, Andreea and Sarlos, Tamas and Hawkins, Peter and Davis, Jared and Mohiuddin, Afroz and Kaiser, Lukasz and Belanger, David and Colwell, Lucy and Weller, Adrian},
  booktitle={ICLR},
  year={2021}
}

@article{linformer,
  title={Linformer: Self-attention with linear complexity},
  author={Wang, Sinong and Li, Belinda Z and Khabsa, Madian and Fang, Hanwen and Ma, Hao},
  journal={arXiv preprint arXiv:2006.04768},
  year={2020}
}

@inproceedings{polaformer,
  title={{PolaFormer}: Polarity-aware Linear Attention for Vision Transformers},
  author={Meng, Weikang and Luo, Yadan and Li, Xin and Jiang, Dongmei and Zhang, Zheng},
  booktitle={ICLR},
  year={2025}
}

@inproceedings{nystromformer,
  title={{Nystr{\"o}mformer}: A {Nystr{\"o}m}-based algorithm for approximating self-attention},
  author={Xiong, Yunyang and Zeng, Zhanpeng and Chakraborty, Rudrasis and Tan, Mingxing and Fung, Glenn and Li, Yin and Singh, Vikas},
  booktitle={AAAI},
  year={2021}
}

@inproceedings{agent_attention,
  title={Agent Attention: On the Integration of Softmax and Linear Attention},
  author={Han, Dongchen and Ye, Tianzhu and Han, Yizeng and Xia, Zhuofan and Pan, Siyuan and Wan, Pengfei and Song, Shiji and Huang, Gao},
  booktitle={ECCV},
  year={2024}
}

@inproceedings{flatten_transformer,
  title={{FLatten} Transformer: Vision Transformer using Focused Linear Attention},
  author={Han, Dongchen and Pan, Xuran and Han, Yizeng and Song, Shiji and Huang, Gao},
  booktitle={ICCV},
  year={2023}
}

@inproceedings{inline_attention,
  title={Bridging the Divide: Reconsidering Softmax and Linear Attention},
  author={Han, Dongchen and Pu, Yifan and Xia, Zhuofan and Han, Yizeng and Pan, Xuran and Li, Xiu and Lu, Jiwen and Song, Shiji and Huang, Gao},
  booktitle={NeurIPS},
  year={2024}
}

@inproceedings{dat,
  title={Vision transformer with deformable attention},
  author={Xia, Zhuofan and Pan, Xuran and Song, Shiji and Li, Li Erran and Huang, Gao},
  booktitle={CVPR},
  year={2022}
}

@inproceedings{biformer,
  title={{BiFormer}: Vision Transformer with Bi-Level Routing Attention},
  author={Zhu, Lei and Wang, Xinjiang and Ke, Zhanghan and Zhang, Wayne and Lau, Rynson W. H.},
  booktitle={CVPR},
  year={2023}
}

@inproceedings{slot_attention,
  title={Object-centric learning with slot attention},
  author={Locatello, Francesco and Weissenborn, Dirk and Unterthiner, Thomas and Mahendran, Aravindh and Heigold, Georg and Uszkoreit, Jakob and Dosovitskiy, Alexey and Kipf, Thomas},
  booktitle={NeurIPS},
  year={2020}
}

@inproceedings{groupvit,
  title={{GroupViT}: Semantic Segmentation Emerges From Text Supervision},
  author={Xu, Jiarui and De Mello, Shalini and Liu, Sifei and Byeon, Wonmin and Breuel, Thomas and Kautz, Jan and Wang, Xiaolong},
  booktitle={CVPR},
  year={2022}
}

@inproceedings{perceiver_io,
  title={{Perceiver IO}: A General Architecture for Structured Inputs \& Outputs},
  author={Jaegle, Andrew and Borgeaud, Sebastian and Alayrac, Jean-Baptiste and Doersch, Carl and Ionescu, Catalin and others},
  booktitle={ICLR},
  year={2022}
}

@inproceedings{kmax_deeplab,
  title={k-means Mask Transformer},
  author={Yu, Qihang and Wang, Huiyu and Qiao, Siyuan and Collins, Maxwell and Zhu, Yukun and Adam, Hartwig and Yuille, Alan and Chen, Liang-Chieh},
  booktitle={ECCV},
  year={2022}
}

@inproceedings{mask_rcnn,
  title={Mask {R-CNN}},
  author={He, Kaiming and Gkioxari, Georgia and Doll{\'a}r, Piotr and Girshick, Ross},
  booktitle={ICCV},
  year={2017}
}

@inproceedings{semantic_fpn,
  title={Panoptic Feature Pyramid Networks},
  author={Kirillov, Alexander and Girshick, Ross and He, Kaiming and Doll{\'a}r, Piotr},
  booktitle={CVPR},
  year={2019}
}

@inproceedings{upernet,
  title={Unified Perceptual Parsing for Scene Understanding},
  author={Xiao, Tete and Liu, Yingcheng and Zhou, Bolei and Jiang, Yuning and Sun, Jian},
  booktitle={ECCV},
  year={2018}
}

@inproceedings{imagenet,
  title={{ImageNet}: A Large-Scale Hierarchical Image Database},
  author={Deng, Jia and Dong, Wei and Socher, Richard and Li, Li-Jia and Li, Kai and Fei-Fei, Li},
  booktitle={CVPR},
  year={2009}
}

@inproceedings{coco,
  title={Microsoft {COCO}: Common Objects in Context},
  author={Lin, Tsung-Yi and Maire, Michael and Belongie, Serge and Hays, James and Perona, Pietro and Ramanan, Deva and Doll{\'a}r, Piotr and Zitnick, C. Lawrence},
  booktitle={ECCV},
  year={2014}
}

@article{ade20k,
  title={Semantic Understanding of Scenes Through the {ADE20K} Dataset},
  author={Zhou, Bolei and Zhao, Hang and Puig, Xavier and Xiao, Tete and Fidler, Sanja and Barriuso, Adela and Torralba, Antonio},
  journal={IJCV},
  year={2019}
}

@inproceedings{adamw,
  title={Decoupled Weight Decay Regularization},
  author={Loshchilov, Ilya and Hutter, Frank},
  booktitle={ICLR},
  year={2019}
}

@inproceedings{ ca_learners,
  title={Vision transformers are circulant attention learners},
  author={Han, Dongchen and Li, Tianyu and Wang, Ziyi and Huang, Gao},
  booktitle={AAAI},
  year={2026}
}

@inproceedings{mobiatten,
  title={Mobile attention: mobile-friendly linear-attention for vision transformers},
  author={Yao, Zhiyu and Wang, Jian and Wu, Haixu and Wang, Jingdong and Long, Mingsheng},
  booktitle={ICML},
  year={2024}
}

@article{wu2025lrformer,
  title={Low-Resolution Self-Attention for Semantic Segmentation},
  author={Wu, Yu-Huan and Zhang, Shi-Chen and Liu, Yun and Zhang, Le and
          Zhan, Xin and Zhou, Daquan and Feng, Jiashi and Cheng, Ming-Ming
          and Zhen, Liangli},
  journal={IEEE Transactions on Pattern Analysis and Machine Intelligence},
  volume={47},
  number={9},
  pages={8180--8192},
  year={2025},
  doi={10.1109/TPAMI.2025.3577035}
}
\bibliographystyle{iclr2027_conference}
\clearpage

\appendix
\raggedbottom
\numberwithin{figure}{section}
\numberwithin{table}{section}
\begin{center}
    {\Large\bfseries Appendix}
\end{center}
\vspace{0.5em}
\section{Implementation and Model Architectures}

\subsection{Implementation of RPAttention}
\label{app:implementation}

We provide a compact PyTorch-style implementation of the core
\textit{Gather--Interact--Distribute} computation below, omitting standard module initialization for clarity.
\begin{lstlisting}[style=rpattenpython]
def rpattention_forward(x, H, W):
    B, N, C = x.shape
    qkv = self.qkv(x).reshape(B, N, 3, self.num_heads, self.head_dim).permute(2, 0, 3, 1, 4)
    q, k, v = qkv.unbind(0)

    # Local depth-wise convolution branch
    v_img = v.transpose(1, 2).reshape(B, N, C)
    v_img = v_img.transpose(1, 2).reshape(B, C, H, W)
    local = self.dwc(v_img).flatten(2).transpose(1, 2)

    # Gather: tokens -> representatives
    A = F.softmax(self.gather_proj(k), dim=-1)
    A = A / A.sum(dim=2, keepdim=True).clamp_min(1e-6)
    gather = A.transpose(-1, -2)

    K_l = self.latent_k_norm(gather @ k)
    V_l = self.latent_v_norm(gather @ v)

    # Interact: self-attention among representatives
    lqkv = self.latent_qkv(V_l).reshape(
        B, self.num_heads, self.num_latents, 3, self.head_dim
    ).permute(3, 0, 1, 2, 4)
    lq, lk, lv = lqkv.unbind(0)

    attn_l = (lq @ lk.transpose(-2, -1)) * self.scale
    V_l = V_l + F.softmax(attn_l, dim=-1) @ lv

    # Distribute: representatives -> tokens
    attn = (q @ K_l.transpose(-2, -1)) * self.scale
    global_out = F.softmax(attn, dim=-1) @ V_l
    global_out = global_out.transpose(1, 2).reshape(B, N, C)

    return self.out_proj(global_out + local)
\end{lstlisting}

\subsection{Model Architectures}
\label{app:model_architectures}
We summarize the architectures of the RPAttention models used in the
main experiments. Spatial resolutions are reported for a
$224\times224$ input.
For each stage, we list the embedding dimension, number of attention
heads, number of representatives per head, and block depth.
The notation $\times L$ denotes $L$ repeated blocks, and LN denotes
LayerNorm.
RPAttention replaces the attention module in every block,
while the remaining backbone components are preserved.
\begin{table}[H]
\centering
\caption{Architectures of RPAttention-DeiT models.}
\label{tab:arch_rpatten_deit}

\begingroup
\small
\setlength{\tabcolsep}{7pt}
\renewcommand{\arraystretch}{1.15}

\begin{tabular}{cc|cc}
\toprule
\textbf{Stage}
& \textbf{Resolution}
& \textbf{RPAttention-DeiT-T}
& \textbf{RPAttention-DeiT-S} \\
\midrule

\multicolumn{2}{c|}{Patch embedding}
& \multicolumn{2}{c}{Patch $16\times16$, stride $16$} \\
\midrule

res1
& $14\times14$
& $\left[\begin{array}{c}
\mathrm{dim}\ 192\\
\mathrm{head}\ 3\\
\mathrm{rep.}\ 49
\end{array}\right]\times12$
& $\left[\begin{array}{c}
\mathrm{dim}\ 384\\
\mathrm{head}\ 6\\
\mathrm{rep.}\ 49
\end{array}\right]\times12$
\\

\bottomrule
\end{tabular}

\endgroup
\end{table}

\begin{table}[H]
\centering
\caption{Architectures of RPAttention-PVT models.}
\label{tab:arch_rpatten_pvt}

\begingroup
\small
\setlength{\tabcolsep}{5pt}
\renewcommand{\arraystretch}{1.18}

\resizebox{\linewidth}{!}{%
\begin{tabular}{cc|cccc}
\toprule
\textbf{Stage}
& \textbf{Resolution}
& \textbf{RPAttention-PVT-T}
& \textbf{RPAttention-PVT-S}
& \textbf{RPAttention-PVT-M}
& \textbf{RPAttention-PVT-L} \\
\midrule

\multicolumn{2}{c|}{Patch embedding}
& \multicolumn{4}{c}{Conv $4\times4$, stride $4$, 64, LN} \\
\midrule

res1
& $56\times56$
& $\left[\begin{array}{c}
\mathrm{dim}\ 64\\
\mathrm{head}\ 1\\
\mathrm{rep.}\ 96
\end{array}\right]\times2$
& $\left[\begin{array}{c}
\mathrm{dim}\ 64\\
\mathrm{head}\ 1\\
\mathrm{rep.}\ 96
\end{array}\right]\times3$
& $\left[\begin{array}{c}
\mathrm{dim}\ 64\\
\mathrm{head}\ 1\\
\mathrm{rep.}\ 96
\end{array}\right]\times3$
& $\left[\begin{array}{c}
\mathrm{dim}\ 64\\
\mathrm{head}\ 1\\
\mathrm{rep.}\ 96
\end{array}\right]\times3$
\\
\midrule

\multicolumn{2}{c|}{Patch embedding}
& \multicolumn{4}{c}{Conv $2\times2$, stride $2$, 128, LN} \\
\midrule

res2
& $28\times28$
& $\left[\begin{array}{c}
\mathrm{dim}\ 128\\
\mathrm{head}\ 2\\
\mathrm{rep.}\ 64
\end{array}\right]\times2$
& $\left[\begin{array}{c}
\mathrm{dim}\ 128\\
\mathrm{head}\ 2\\
\mathrm{rep.}\ 64
\end{array}\right]\times4$
& $\left[\begin{array}{c}
\mathrm{dim}\ 128\\
\mathrm{head}\ 2\\
\mathrm{rep.}\ 64
\end{array}\right]\times4$
& $\left[\begin{array}{c}
\mathrm{dim}\ 128\\
\mathrm{head}\ 2\\
\mathrm{rep.}\ 64
\end{array}\right]\times8$
\\
\midrule

\multicolumn{2}{c|}{Patch embedding}
& \multicolumn{4}{c}{Conv $2\times2$, stride $2$, 320, LN} \\
\midrule

res3
& $14\times14$
& $\left[\begin{array}{c}
\mathrm{dim}\ 320\\
\mathrm{head}\ 5\\
\mathrm{rep.}\ 49
\end{array}\right]\times2$
& $\left[\begin{array}{c}
\mathrm{dim}\ 320\\
\mathrm{head}\ 5\\
\mathrm{rep.}\ 49
\end{array}\right]\times6$
& $\left[\begin{array}{c}
\mathrm{dim}\ 320\\
\mathrm{head}\ 5\\
\mathrm{rep.}\ 49
\end{array}\right]\times18$
& $\left[\begin{array}{c}
\mathrm{dim}\ 320\\
\mathrm{head}\ 5\\
\mathrm{rep.}\ 49
\end{array}\right]\times27$
\\
\midrule

\multicolumn{2}{c|}{Patch embedding}
& \multicolumn{4}{c}{Conv $2\times2$, stride $2$, 512, LN} \\
\midrule

res4
& $7\times7$
& $\left[\begin{array}{c}
\mathrm{dim}\ 512\\
\mathrm{head}\ 8\\
\mathrm{rep.}\ 49
\end{array}\right]\times2$
& $\left[\begin{array}{c}
\mathrm{dim}\ 512\\
\mathrm{head}\ 8\\
\mathrm{rep.}\ 49
\end{array}\right]\times3$
& $\left[\begin{array}{c}
\mathrm{dim}\ 512\\
\mathrm{head}\ 8\\
\mathrm{rep.}\ 49
\end{array}\right]\times3$
& $\left[\begin{array}{c}
\mathrm{dim}\ 512\\
\mathrm{head}\ 8\\
\mathrm{rep.}\ 49
\end{array}\right]\times3$
\\

\bottomrule
\end{tabular}%
}

\endgroup
\end{table}
\begin{table}[H]
\centering
\caption{Architectures of RPAttention-PVTv2 models.}
\label{tab:arch_rpatten_pvtv2}

\begingroup
\small
\setlength{\tabcolsep}{5pt}
\renewcommand{\arraystretch}{1.18}

\resizebox{\linewidth}{!}{%
\begin{tabular}{cc|ccc}
\toprule
\textbf{Stage}
& \textbf{Resolution}
& \textbf{RPAttention-PVTv2-B0}
& \textbf{RPAttention-PVTv2-B1}
& \textbf{RPAttention-PVTv2-B2} \\
\midrule

res1
& $56\times56$
& $\left[\begin{array}{c}
\mathrm{dim}\ 32\\
\mathrm{head}\ 1\\
\mathrm{rep.}\ 96
\end{array}\right]\times2$
& $\left[\begin{array}{c}
\mathrm{dim}\ 64\\
\mathrm{head}\ 1\\
\mathrm{rep.}\ 96
\end{array}\right]\times2$
& $\left[\begin{array}{c}
\mathrm{dim}\ 64\\
\mathrm{head}\ 1\\
\mathrm{rep.}\ 96
\end{array}\right]\times3$
\\
\midrule

res2
& $28\times28$
& $\left[\begin{array}{c}
\mathrm{dim}\ 64\\
\mathrm{head}\ 2\\
\mathrm{rep.}\ 64
\end{array}\right]\times2$
& $\left[\begin{array}{c}
\mathrm{dim}\ 128\\
\mathrm{head}\ 2\\
\mathrm{rep.}\ 64
\end{array}\right]\times2$
& $\left[\begin{array}{c}
\mathrm{dim}\ 128\\
\mathrm{head}\ 2\\
\mathrm{rep.}\ 64
\end{array}\right]\times4$
\\
\midrule

res3
& $14\times14$
& $\left[\begin{array}{c}
\mathrm{dim}\ 160\\
\mathrm{head}\ 5\\
\mathrm{rep.}\ 49
\end{array}\right]\times2$
& $\left[\begin{array}{c}
\mathrm{dim}\ 320\\
\mathrm{head}\ 5\\
\mathrm{rep.}\ 49
\end{array}\right]\times2$
& $\left[\begin{array}{c}
\mathrm{dim}\ 320\\
\mathrm{head}\ 5\\
\mathrm{rep.}\ 49
\end{array}\right]\times6$
\\
\midrule

res4
& $7\times7$
& $\left[\begin{array}{c}
\mathrm{dim}\ 256\\
\mathrm{head}\ 8\\
\mathrm{rep.}\ 49
\end{array}\right]\times2$
& $\left[\begin{array}{c}
\mathrm{dim}\ 512\\
\mathrm{head}\ 8\\
\mathrm{rep.}\ 49
\end{array}\right]\times2$
& $\left[\begin{array}{c}
\mathrm{dim}\ 512\\
\mathrm{head}\ 8\\
\mathrm{rep.}\ 49
\end{array}\right]\times3$
\\

\bottomrule
\end{tabular}%
}

\endgroup
\end{table}

\begin{table}[H]
\centering
\caption{Architectures of RPAttention-Swin models.}
\label{tab:arch_rpatten_swin}

\begingroup
\small
\setlength{\tabcolsep}{6pt}
\renewcommand{\arraystretch}{1.18}

\begin{tabular}{cc|cc}
\toprule
\textbf{Stage}
& \textbf{Resolution}
& \textbf{RPAttention-Swin-T}
& \textbf{RPAttention-Swin-S} \\
\midrule

\multicolumn{2}{c|}{Patch embedding}
& \multicolumn{2}{c}{Patch $4\times4$, stride $4$, 96, LN} \\
\midrule

res1
& $56\times56$
& $\left[\begin{array}{c}
\mathrm{dim}\ 96\\
\mathrm{head}\ 3\\
\mathrm{rep.}\ 96
\end{array}\right]\times2$
& $\left[\begin{array}{c}
\mathrm{dim}\ 96\\
\mathrm{head}\ 3\\
\mathrm{rep.}\ 96
\end{array}\right]\times2$
\\
\midrule

\multicolumn{2}{c|}{Patch merging}
& \multicolumn{2}{c}{$2\times2$, 192, LN} \\
\midrule

res2
& $28\times28$
& $\left[\begin{array}{c}
\mathrm{dim}\ 192\\
\mathrm{head}\ 6\\
\mathrm{rep.}\ 64
\end{array}\right]\times2$
& $\left[\begin{array}{c}
\mathrm{dim}\ 192\\
\mathrm{head}\ 6\\
\mathrm{rep.}\ 64
\end{array}\right]\times2$
\\
\midrule

\multicolumn{2}{c|}{Patch merging}
& \multicolumn{2}{c}{$2\times2$, 384, LN} \\
\midrule

res3
& $14\times14$
& $\left[\begin{array}{c}
\mathrm{dim}\ 384\\
\mathrm{head}\ 12\\
\mathrm{rep.}\ 49
\end{array}\right]\times6$
& $\left[\begin{array}{c}
\mathrm{dim}\ 384\\
\mathrm{head}\ 12\\
\mathrm{rep.}\ 49
\end{array}\right]\times18$
\\
\midrule

\multicolumn{2}{c|}{Patch merging}
& \multicolumn{2}{c}{$2\times2$, 768, LN} \\
\midrule

res4
& $7\times7$
& $\left[\begin{array}{c}
\mathrm{dim}\ 768\\
\mathrm{head}\ 24\\
\mathrm{rep.}\ 49
\end{array}\right]\times2$
& $\left[\begin{array}{c}
\mathrm{dim}\ 768\\
\mathrm{head}\ 24\\
\mathrm{rep.}\ 49
\end{array}\right]\times2$
\\

\bottomrule
\end{tabular}

\endgroup
\end{table}

\section{Training Details}
\label{app:training}

\subsection{ImageNet-1K Classification}
\begin{table}[H]
\centering
\small
\caption{Training configurations for ImageNet-1K classification.}
\label{tab:imagenet_training_recipe}
\begin{tabular}{lc}
\toprule
\textbf{Setting} & \textbf{Value} \\
\midrule
Input resolution & $224 \times 224$ \\
Training length & 300 epochs \\
Optimizer & AdamW \\
Weight decay & 0.05 \\
LR schedule & Cosine decay to $10^{-5}$ \\
Warm-up & 20 epochs, linear from $10^{-6}$ \\
RandAugment & rand-m9-mstd0.5-inc1 \\
Mixup / CutMix & 0.8 / 1.0 \\
Label smoothing & 0.1 \\
Random erasing & 0.25 \\
Random resized crop & scale $(0.08,1.0)$, ratio $(0.75,1.33)$ \\
Horizontal flip & 0.5 \\
Drop-path rate & Model-specific (see below) \\
EMA & No \\
Default seed & 42 \\
\midrule
Hardware & TPU v4-32 \\
Local batch size & 256 per process \\
Global batch size & 4096 \\
Peak learning rate & $4 \times 10^{-3}$ \\
\bottomrule
\end{tabular}
\end{table}
All RPAttention classification models reported in the main tables
are trained on a TPU v4-32 with a global batch size of 4096
and a peak learning rate of $4\times10^{-3}$.
The drop-path rates are 0.0/0.1 for DeiT-T/S,
0.0/0.1/0.3/0.3 for PVT-T/S/M/L,
0.0/0.1/0.2 for PVTv2-B0/B1/B2,
and 0.2/0.4 for Swin-T/S.
The remaining settings in Table~\ref{tab:imagenet_training_recipe}
are shared across these models.

\subsection{Dense Prediction}

\begin{table}[H]
\centering
\small
\caption{Training configurations for COCO object detection and instance segmentation.}
\label{tab:coco_training}
\renewcommand{\arraystretch}{1.12}
\begin{tabular}{@{}lcc@{}}
\toprule
\textbf{Setting} & \textbf{PVT / PVTv2} & \textbf{Swin} \\
\midrule
Framework
& \multicolumn{2}{c}{Mask R-CNN} \\
Training length
& \multicolumn{2}{c}{$1\times$ (12 epochs)} \\
Optimizer
& \multicolumn{2}{c}{AdamW} \\
Weight decay
& $1\times10^{-4}$
& $5\times10^{-2}$ \\
Global batch size
& \multicolumn{2}{c}{16} \\
Learning rate
& $2\times10^{-4}$
& $1\times10^{-4}$ \\
LR schedule
& \multicolumn{2}{c}{Step decay at epochs 8 and 11} \\
Warm-up
& \multicolumn{2}{c}{500 iterations, linear, ratio $10^{-3}$} \\

Test resize scale ($W\times H$)
& \multicolumn{2}{c}{$1333\times800$, keep ratio} \\

Test augmentation
& \multicolumn{2}{c}{Single scale, no flipping} \\
\bottomrule
\end{tabular}
\end{table}
For PVT and PVTv2, COCO training uses a resize scale of
$(1333,800)$ with aspect ratio preserved.
For Swin, training uses multiscale resizing with short-side lengths
from 480 to 800 in steps of 32 and a long-side limit of 1333,
with an additional resize--crop--resize augmentation branch.
Both settings use random horizontal flipping with probability 0.5.

\begin{table}[H]
\centering
\small
\caption{Training configurations for ADE20K semantic segmentation.}
\label{tab:ade20k_training}
\renewcommand{\arraystretch}{1.12}
\begin{tabular}{@{}lcc@{}}
\toprule
\textbf{Setting} & \textbf{PVT / PVTv2} & \textbf{Swin} \\
\midrule
Segmentation head
& Semantic FPN~\citep{semantic_fpn}
& UPerNet~\citep{upernet} \\

Optimizer
& \multicolumn{2}{c}{AdamW} \\

Global batch size
& 16 / 32 
& 16 \\

Training crop
& \multicolumn{2}{c}{$512\times512$} \\

Resize scale range
& \multicolumn{2}{c}{$[0.5,2.0]$} \\

Horizontal flip probability
& \multicolumn{2}{c}{0.5} \\

Photometric distortion
& \multicolumn{2}{c}{Yes} \\

Test resize scale ($W\times H$)
& \multicolumn{2}{c}{$2048\times512$, keep ratio} \\

Test augmentation
& \multicolumn{2}{c}{Single scale, no flipping} \\

\midrule
Training iterations
& 40,000
& 160,000 \\

Learning rate
& $2\times10^{-4}$
& $6\times10^{-5}$ \\

Weight decay
& $1\times10^{-4}$
& $1\times10^{-2}$ \\

LR schedule
& Poly ($p=0.9$)
& Poly ($p=1.0$) \\

Minimum learning rate
& \multicolumn{2}{c}{0} \\

Warm-up
& None
& 1,500 iterations, linear \\

\bottomrule
\end{tabular}
\end{table}

\noindent
All RPAttention backbones used for downstream tasks are initialized
from their corresponding ImageNet-1K checkpoints.
Tables~\ref{tab:coco_training} and~\ref{tab:ade20k_training}
provide reference training configurations for COCO and ADE20K,
respectively.

\section{Practical Efficiency}
\label{sec:efficiency_protocol}

\subsection{Measurement protocol}
All practical-efficiency measurements in
Figure~\ref{fig:efficiency_scaling} and this section are conducted on
a single otherwise idle NVIDIA RTX~3090 using eager PyTorch 2.7.1,
CUDA 11.8, cuDNN 9.1, and timm 0.4.12.
We use AMP/FP16 inference with TF32 disabled,
\texttt{model.eval()}, and \texttt{torch.inference\_mode()},
without \texttt{torch.compile} or inference-specific operator fusion.
CPU intra-operation parallelism is restricted to one thread, while
cuDNN benchmarking is enabled.

Each model--resolution--batch configuration is evaluated in a fresh
process.
Input tensors are generated in advance and remain resident on the GPU;
data loading, preprocessing, host-to-device transfer, and
task-specific post-processing are excluded.
The models are randomly initialized for execution-cost measurement.
This does not alter the measured execution graph because all compared
methods use dense computation without input-dependent token pruning
or conditional execution.

Latency is measured as synchronized wall-clock model-forward latency
with batch size 1.
Throughput uses batch sizes 64, 16, 8, and 4 at resolutions
$224^2$, $512^2$, $768^2$, and $1024^2$, respectively, with the
batch size fixed across methods at each resolution.
Peak allocated memory is measured after warm-up and after resetting
the PyTorch peak-memory counter.
Figures report memory at batch size 1;
Table~\ref{tab:pvt_t_resolution_absolute} additionally reports memory
at the batch size used for throughput measurement.
Each memory value is measured once per configuration.
Except for the rectangular-input measurements in
Section~\ref{sec:rectangular_proxy}, each runtime measurement consists
of five rounds, each with 50 warm-up forwards followed by 200 timed
forwards. For these five-round measurements, we report the mean across the round means;
Table~\ref{tab:pvt_t_resolution_absolute} also reports their
standard deviation.
The maximum coefficient of variation among the throughput
measurements in Figure~\ref{fig:efficiency_scaling} is below
$0.36\%$.

For PVT-based models, each network is instantiated at its target
resolution so that its absolute positional embeddings match the
input resolution; all methods follow the same construction rule.
Therefore, the high-resolution sweep measures execution-cost scaling
rather than high-resolution classification accuracy.
PVT, Agent Attention, FLatten, and PolaFormer use their official
implementations.

\begin{figure*}[t]
    \centering
    \includegraphics[
        width=\textwidth
    ]{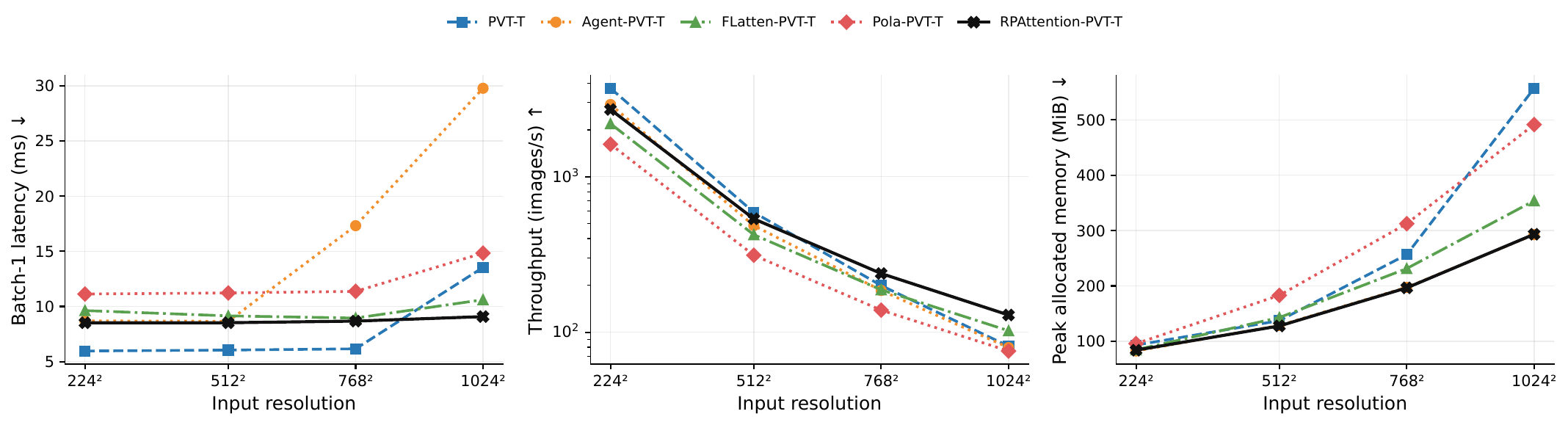}
    \caption{
        Absolute execution-cost measurements for PVT-T and
        representative efficient-attention variants on a single
        NVIDIA RTX~3090.
        Left: synchronized batch-1 model-forward latency.
        Middle: matched-batch throughput, using batch sizes
        64, 16, 8, and 4 at $224^2$, $512^2$, $768^2$, and
        $1024^2$, respectively.
        Right: batch-1 peak allocated GPU memory.
        All methods use eager AMP/FP16 inference under the same
        software and measurement protocol.
    }
    \label{fig:absolute_efficiency}
\end{figure*}

\subsection{Absolute execution cost}
Figure~\ref{fig:absolute_efficiency} presents absolute execution
costs across input resolutions, complementing the normalized results
in the top panels of Figure~\ref{fig:efficiency_scaling}.
Table~\ref{tab:pvt_t_resolution_absolute} provides the corresponding
numerical results and runtime variability, together with peak memory
at the throughput batch size.
At $224^2$, RPAttention-PVT-T incurs additional execution overhead
relative to PVT-T, with a batch-1 latency of 8.52\,ms versus
5.96\,ms and a throughput of 2706 versus 3706 images/s.
At $768^2$, RPAttention achieves higher throughput than PVT-T,
although its batch-1 latency remains higher.
At $1024^2$, RPAttention reduces batch-1 latency from 13.53\,ms to
9.08\,ms and increases matched-batch throughput from 81.3 to
129.0 images/s.
Its batch-1 peak allocated memory decreases from 557.2\,MiB to
293.3\,MiB.

At $1024^2$, RPAttention provides $1.62\times$, $1.26\times$,
and $1.70\times$ the throughput of Agent Attention,
FLatten, and PolaFormer, respectively.
Its peak memory is comparable to Agent Attention, while
remaining lower than FLatten and PolaFormer.
These measurements demonstrate a resolution-dependent crossover
rather than a universal speed advantage at every operating point.
The observed trend is consistent with the fixed-representative
interaction cost $\mathcal{O}(NM+M^2)$ becoming more favorable as
the spatial-token count $N$ increases.
\begin{table}[t]
\centering
\caption{Absolute inference cost of PVT-T and its attention variants
on an RTX 3090 under the protocol above.
Latency uses batch size 1; throughput uses batch size $B$.
Runtime values report mean $\pm$ standard deviation across five rounds.
$M_1$ and $M_B$ denote peak allocated memory at batch sizes 1 and $B$,
respectively, each measured once after warm-up.}
\label{tab:pvt_t_resolution_absolute}
\normalsize
\setlength{\tabcolsep}{3pt}
\renewcommand{\arraystretch}{1.08}
\begin{tabular}{@{}llcccrr@{}}
\toprule
\textbf{Input}
& \textbf{Method}
& $\boldsymbol{B}$
& \textbf{Latency}$\downarrow$
& \textbf{Throughput}$\uparrow$
& $\boldsymbol{M_1}\downarrow$
& $\boldsymbol{M_B}\downarrow$ \\
& & & (ms) & (images/s) & (MiB) & (MiB) \\
\midrule
$224^2$ & PVT-T & 64
& $5.961 \pm 0.015$ & $3706.11 \pm 6.36$ & 93.0 & 662.0 \\
& Agent Attention & 64
& $8.690 \pm 0.032$ & $2912.82 \pm 6.53$ & 82.3 & 652.0 \\
& FLatten & 64
& $9.625 \pm 0.053$ & $2192.96 \pm 2.02$ & 84.5 & 654.2 \\
& PolaFormer & 64
& $11.127 \pm 0.008$ & $1619.20 \pm 0.95$ & 95.6 & 1046.7 \\
\rowcolor{verylightgray}
& RPAttention & 64
& $8.523 \pm 0.046$ & $2706.24 \pm 1.57$ & 83.8 & 652.7 \\
\midrule
$512^2$ & PVT-T & 16
& $6.044 \pm 0.019$ & $590.28 \pm 0.63$ & 137.1 & 841.5 \\
& Agent Attention & 16
& $8.655 \pm 0.031$ & $484.63 \pm 0.96$ & 128.6 & 832.6 \\
& FLatten & 16
& $9.144 \pm 0.041$ & $423.61 \pm 0.60$ & 142.8 & 848.5 \\
& PolaFormer & 16
& $11.224 \pm 0.038$ & $312.53 \pm 0.18$ & 183.0 & 1350.1 \\
\rowcolor{verylightgray}
& RPAttention & 16
& $8.522 \pm 0.032$ & $535.90 \pm 0.28$ & 127.6 & 832.7 \\
\midrule
$768^2$ & PVT-T & 8
& $6.158 \pm 0.011$ & $199.49 \pm 0.22$ & 256.5 & 1345.1 \\
& Agent Attention & 8
& $17.319 \pm 0.008$ & $185.51 \pm 0.13$ & 197.4 & 938.9 \\
& FLatten & 8
& $8.946 \pm 0.038$ & $187.24 \pm 0.09$ & 231.3 & 971.0 \\
& PolaFormer & 8
& $11.365 \pm 0.045$ & $138.56 \pm 0.09$ & 312.4 & 1536.1 \\
\rowcolor{verylightgray}
& RPAttention & 8
& $8.672 \pm 0.062$ & $238.42 \pm 0.15$ & 196.4 & 938.5 \\
\midrule
$1024^2$ & PVT-T & 4
& $13.535 \pm 0.004$ & $81.30 \pm 0.04$ & 557.2 & 1891.3 \\
& Agent Attention & 4
& $29.775 \pm 0.004$ & $79.58 \pm 0.02$ & 292.6 & 856.5 \\
& FLatten & 4
& $10.608 \pm 0.021$ & $102.16 \pm 0.06$ & 353.9 & 917.3 \\
& PolaFormer & 4
& $14.834 \pm 0.004$ & $75.67 \pm 0.05$ & 491.7 & 1418.7 \\
\rowcolor{verylightgray}
& RPAttention & 4
& $9.077 \pm 0.021$ & $128.95 \pm 0.07$ & 293.3 & 855.9 \\
\bottomrule
\end{tabular}
\end{table}

\begin{table}[t]
    \centering
    \caption{Relative changes versus the corresponding PVT backbone.
    Higher throughput and lower latency or memory are better.}
    \label{tab:scale_efficiency}
    \small
    \setlength{\tabcolsep}{4pt}
    \begin{tabular}{lrrrrrr}
        \toprule
        & \multicolumn{4}{c}{Throughput change}
        & \multicolumn{2}{c}{Change at $1024^2$} \\
        \cmidrule(lr){2-5}
        \cmidrule(lr){6-7}
        Scale
        & $224^2$
        & $512^2$
        & $768^2$
        & $1024^2$
        & Latency
        & Memory \\
        \midrule
        Tiny
        & $-27.0\%$
        & $-9.2\%$
        & $+19.5\%$
        & $+58.6\%$
        & $-32.9\%$
        & $-47.4\%$ \\
        Small
        & $-27.7\%$
        & $-8.1\%$
        & $+21.9\%$
        & $+62.8\%$
        & $-33.6\%$
        & $-44.5\%$ \\
        Medium
        & $-28.3\%$
        & $-6.2\%$
        & $+24.7\%$
        & $+65.4\%$
        & $-24.0\%$
        & $-41.1\%$ \\
        \bottomrule
    \end{tabular}
\end{table}

\subsection{Scaling across model sizes}
Table~\ref{tab:scale_efficiency} extends the resolution sweep from
PVT-Tiny to PVT-Small and PVT-Medium.
All three model sizes exhibit the same matched-batch throughput
crossover:
RPAttention is slower at $224^2$ and $512^2$, becomes faster at
$768^2$, and provides a $58.6\%$--$65.4\%$ throughput improvement
at $1024^2$.
Batch-1 latency and peak memory also decrease consistently at
$1024^2$.
The observed crossover is therefore not specific to the Tiny model.

\subsection{Interaction-range scaling}
Figure~\ref{fig:efficiency_scaling} (bottom) varies Swin-T's window
size while fixing RPAttention-Swin-T to RRRR with window size 56
and $(96,64,49,49)$ representatives.
All models use $224\times224$ inputs, with batch sizes 64 and 1
for throughput and latency, respectively.
RPAttention has lower throughput at window size 7 and comparable
throughput at 14.
At sizes 28 and 56, it uses $17.2\%$ and $43.9\%$ fewer FLOPs
and achieves $1.89\times$ and $3.89\times$ the throughput of Swin-T,
respectively.
This comparison measures execution cost, not accuracy equivalence.

\begingroup
\subsection{Accuracy and Practical Efficiency}
\label{sec:pvt_accuracy_cost}

To complement the accuracy--FLOPs comparison, Figure~\ref{fig:pvt_imagenet_cost}
compares ImageNet-1K Top-1 accuracy with batch-1 latency and peak GPU memory
across PVT scales and attention
variants~\citep{pvt,agent_attention,flatten_transformer,polaformer,inline_attention}.
T, S, M, and L denote Tiny, Small, Medium, and Large, respectively.
Following the measurement protocol in Appendix~\ref{sec:efficiency_protocol},
latency and memory are measured on an RTX~3090 with FP16, batch size 1,
and $224\times224$ inputs. Our main-paper classification training settings
are described in Appendix~\ref{app:training}; competing accuracy results
are taken from the corresponding publications.

Across the evaluated PVT scales, RPAttention improves accuracy and reduces peak memory compared with the original backbone. 
Compared with FLatten,
it achieves higher accuracy with lower latency and memory at all four
scales. Compared with Agent Attention, it runs faster with similar memory
usage, while achieving comparable or higher accuracy.

\subsection{Comparison of Practical Efficiency and Accuracy on Downstream Tasks}
\label{sec:rectangular_proxy}

To complement the accuracy--FLOPs comparison, Figures~\ref{fig:pvt_coco_cost}
and~\ref{fig:pvt_ade20k_cost} and Table~\ref{tab:rectangular_coco_proxy}
compare reported COCO box AP and ADE20K mIoU with FLOPs, latency, and memory.
FLOPs are reported for the detection and segmentation models, while latency
and memory are measured on PVT classification models adapted to
$800\times1280$ and $512\times2048$ inputs, respectively.

Following Section~\ref{sec:pvt_accuracy_cost}, measurements use an RTX~3090,
FP16, and batch size 1.
The rectangular-input latency results are averaged over three rounds,
each containing 20 timed forwards.
For cached Agent Attention, cache construction is excluded from latency
and cached tensors are included in memory. Both settings use the same
reported downstream task metrics.
RPAttention's efficiency advantages become more evident at these resolutions.
Across PVT-T, PVT-S, and PVT-M, it improves task accuracy over the original
backbone while reducing measured latency and peak memory. Compared with
InLine, it achieves higher box AP and mIoU with similar latency and memory.
Compared with Agent Attention, it has lower latency and memory in both
cached and uncached settings; on PVT-M, it also matches box AP and improves
mIoU. Caching accelerates Agent Attention at the cost of substantially
higher memory use.

RPAttention achieves these results while retaining the same representative
counts used for classification, supporting compact, content-adaptive global
communication as spatial resolution increases.

\begin{table}[H]
    \centering
    \caption{Accuracy and practical efficiency on COCO and ADE20K. Bold
    and underlining indicate the best and second-best distinct values within
    each model scale.}
    \label{tab:rectangular_coco_proxy}
    \label{tab:rectangular_ade20k_proxy}

    \begingroup
    \setlength{\tabcolsep}{1.5pt}
\renewcommand{\arraystretch}{0.90}
\resizebox{\linewidth}{!}{%
    \begin{tabular}{@{}lrrrrrrrr@{}}
    \toprule
    \multirow{3}{*}{\textbf{Method}}
    & \multicolumn{4}{c}{\textbf{COCO} ($800\times1280$)}
    & \multicolumn{4}{c}{\textbf{ADE20K} ($512\times2048$)} \\
    \cmidrule(lr){2-5}\cmidrule(l){6-9}
    & AP$^b$ $\uparrow$ & FLOPs $\downarrow$ & Latency $\downarrow$ & Memory $\downarrow$
    & mIoU $\uparrow$ & FLOPs $\downarrow$ & Latency $\downarrow$ & Memory $\downarrow$ \\
    & \multicolumn{1}{c}{(\%)} & \multicolumn{1}{c}{(G)} & \multicolumn{1}{c}{(ms)} & \multicolumn{1}{c}{(MiB)}
    & \multicolumn{1}{c}{(\%)} & \multicolumn{1}{c}{(G)} & \multicolumn{1}{c}{(ms)} & \multicolumn{1}{c}{(MiB)} \\
    \midrule
    PVT-T & 36.7 & 240 & 12.39 & 538.8 & 36.57 & 158 & 13.13 & 557.2 \\
    + Agent (uncached) & \multirow{2}{*}{\textbf{41.4}} & \multirow{2}{*}{230} & 91.83 & 678.7 & \multirow{2}{*}{\textbf{40.18}} & \multirow{2}{*}{147} & 94.04 & 696.8 \\
    + Agent (cached) & & & 15.18 & 923.1 & & & 15.47 & 944.8 \\
    + FLatten & 38.2 & 244 & 10.17 & 348.7 & 37.21 & 169 & 10.31 & 353.9 \\
    + PolaFormer & \underline{40.4} & 216 & 14.26 & 489.6 & 38.30 & 133 & 14.51 & 491.7 \\
    + InLine & 40.2 & \textbf{211} & \underline{8.87} & \underline{290.4} & 39.16 & \textbf{127} & \underline{9.00} & \underline{294.8} \\
    \rowcolor{verylightgray}
    + RPAttention & \underline{40.4} & \underline{215} & \textbf{8.85} & \textbf{288.5} & \underline{40.00} & \underline{128} & \textbf{8.85} & \textbf{293.3} \\
    \midrule
    PVT-S & 40.4 & 305 & 22.28 & 603.3 & 41.95 & 225 & 23.51 & 621.7 \\
    + Agent (uncached) & \multirow{2}{*}{\textbf{44.5}} & \multirow{2}{*}{293} & 178.15 & 730.4 & \multirow{2}{*}{\textbf{44.18}} & \multirow{2}{*}{211} & 182.86 & 749.6 \\
    + Agent (cached) & & & 28.43 & 1700.8 & & & 28.98 & 1742.1 \\
    + PolaFormer & 43.9 & 262 & 25.92 & 599.1 & 41.00 & 179 & 26.37 & 601.9 \\
    + InLine & 43.4 & \textbf{250} & \textbf{15.45} & \underline{345.0} & 42.93 & \textbf{168} & \underline{15.72} & \underline{349.3} \\
    \rowcolor{verylightgray}
    + RPAttention & \underline{44.1} & \underline{259} & \underline{15.72} & \textbf{341.2} & \underline{43.25} & \underline{177} & \textbf{15.70} & \textbf{345.2} \\
    \midrule
    PVT-M & 42.0 & 392 & 33.39 & 716.2 & 42.91 & 315 & 34.96 & 734.5 \\
    + Agent (uncached) & \multirow{2}{*}{\textbf{45.9}} & \multirow{2}{*}{400} & 294.62 & 819.3 & \multirow{2}{*}{44.30} & \multirow{2}{*}{321} & 303.32 & 839.2 \\
    + Agent (cached) & & & 47.54 & 3229.7 & & & 48.47 & 3319.7 \\
    + InLine & \underline{44.0} & \textbf{310} & \textbf{25.44} & \underline{436.9} & \underline{44.59} & \textbf{229} & \textbf{25.23} & \underline{441.2} \\
    \rowcolor{verylightgray}
    + RPAttention & \textbf{45.9} & \underline{322} & \underline{26.36} & \textbf{427.7} & \textbf{45.16} & \underline{241} & \underline{27.57} & \textbf{432.5} \\
    \bottomrule
    \end{tabular}}
    \endgroup
\end{table}
\begin{figure}[p]
    \centering

    \includegraphics[width=0.90\linewidth]{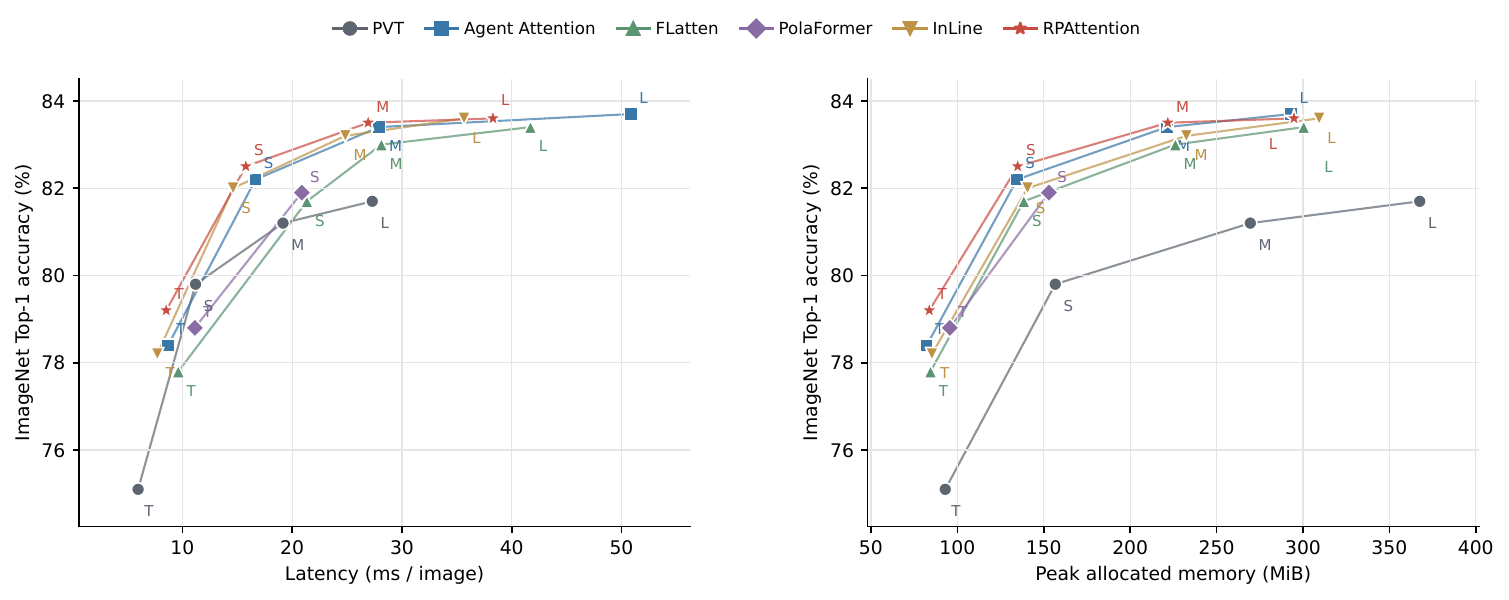}
    \caption{ImageNet-1K Top-1 accuracy versus batch-1 latency and peak
    GPU memory across PVT scales. Latency and memory are measured on an
    RTX~3090 with FP16 and $224\times224$ inputs. Missing model--accuracy
    combinations are omitted.}
    \label{fig:pvt_imagenet_cost}

    \par\vspace{0.4em}

    \includegraphics[width=0.90\linewidth]{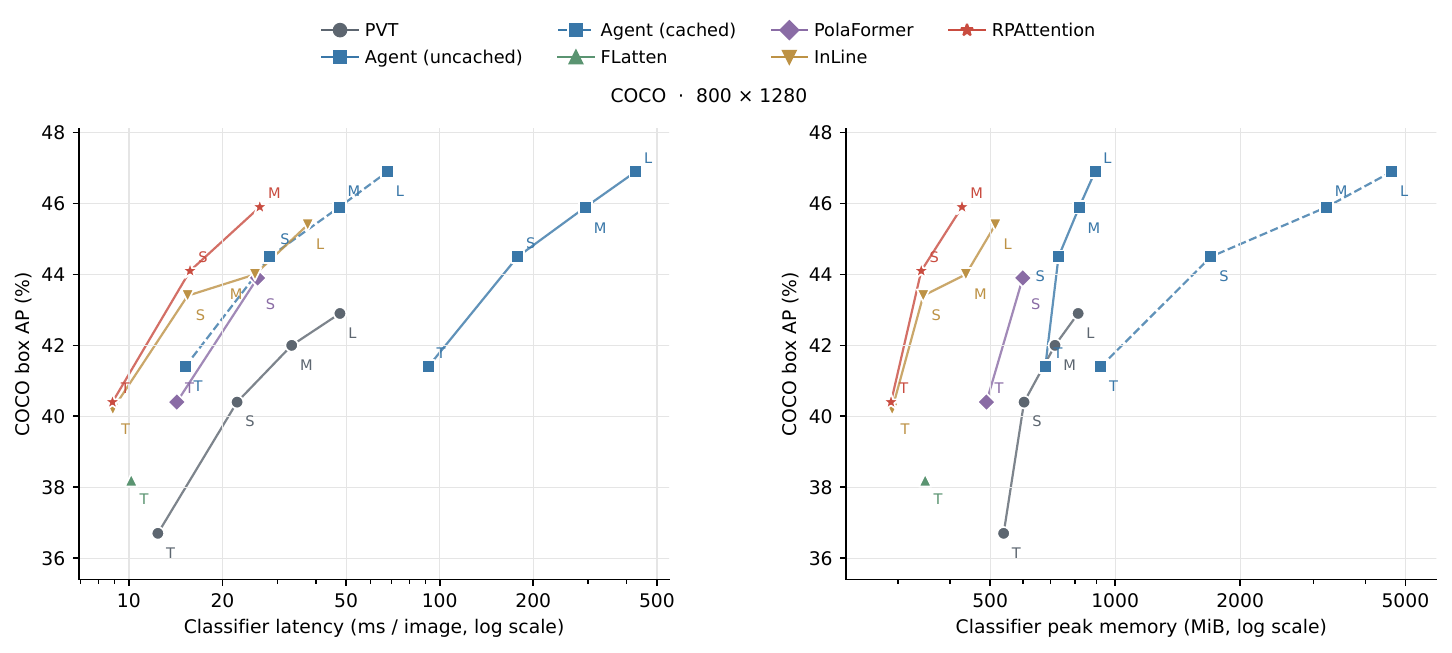}
    \caption{Reported COCO box AP versus rectangular-input classifier
    proxy cost at $800\times1280$. Agent Attention is shown with and without
    fixed-shape caching. The horizontal axes are logarithmic.}
    \label{fig:pvt_coco_cost}

    \par\vspace{0.4em}

    \includegraphics[width=0.90\linewidth]{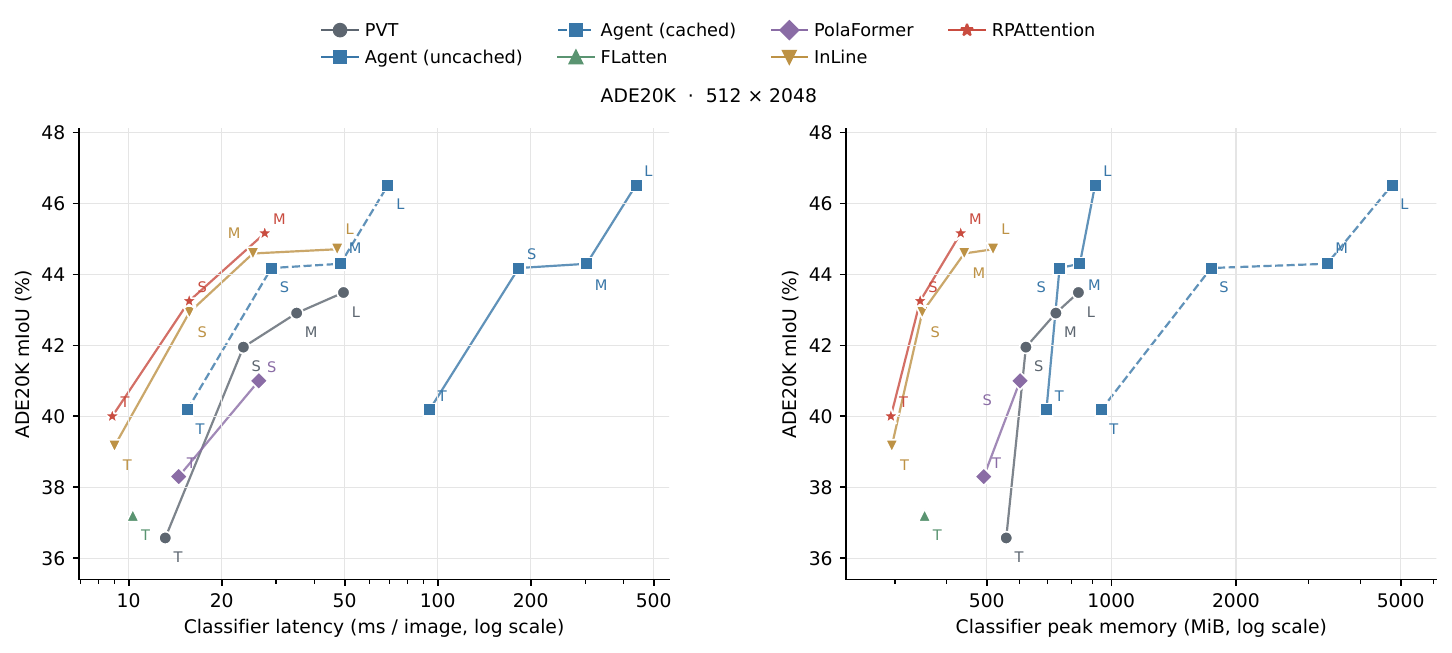}
    \caption{Reported ADE20K mIoU versus rectangular-input classifier
    proxy cost at $512\times2048$. Agent Attention is shown with and without
    fixed-shape caching. The horizontal axes are logarithmic.}
    \label{fig:pvt_ade20k_cost}
\end{figure}
\endgroup
\clearpage

\section{Comparisons with Other Methods}
\label{app:method_comparisons}

We provide additional comparisons with FLatten
\citep{flatten_transformer}, Agent Attention
\citep{agent_attention}, PolaFormer \citep{polaformer},
InLine \citep{inline_attention}, MobiAtt \citep{mobiatten},
and CA \citep{ca_learners}.
Gray shading identifies RPAttention, and boldface indicates
the best task performance within each backbone group.

\subsection{ImageNet-1K Classification}
\label{app:classification_comparisons}
Table~\ref{tab:imagenet_method_comparisons} compares classification
accuracy, parameters, and FLOPs across four backbone families.

\begin{table}[htbp]
\centering
\caption{ImageNet-1K classification comparison with other
efficient-attention methods. Parentheses indicate Top-1 gains
in percentage points over the corresponding backbone.
Best Top-1 accuracy within each group is highlighted in bold.}
\label{tab:imagenet_method_comparisons}
\begingroup
\small
\setlength{\tabcolsep}{2pt}
\setlength{\aboverulesep}{0pt}
\setlength{\belowrulesep}{0pt}
\renewcommand{\arraystretch}{1.12}

\begin{minipage}[t]{0.49\linewidth}
\centering
\begin{tabular}[t]{@{}lccc@{}}
\toprule
\textbf{Method} & \textbf{Params}
& \textbf{FLOPs} & \textbf{Top-1} \\
\midrule
DeiT-T          & 5.7M & 1.2G & 72.2 \\
+ FLatten~\citeyearpar{flatten_transformer} & 6.1M & 1.1G & 74.1 (+1.9) \\
+ Agent~\citeyearpar{agent_attention}         & 6.0M & 1.2G & 74.9 (+2.7) \\
+ PolaFormer~\citeyearpar{polaformer}    & 6.1M & 1.2G & 74.6 (+2.4) \\
+ InLine~\citeyearpar{inline_attention}        & 6.5M & 1.1G & 74.5 (+2.3) \\
\rowcolor{verylightgray}
+ RPAttention   & 5.9M & 1.2G & \textbf{75.4 (+3.2)} \\
\midrule
DeiT-S          & 22.1M & 4.6G & 79.8 \\
+ Agent~\citeyearpar{agent_attention}         & 22.7M & 4.4G & 80.5 (+0.7) \\
+ InLine~\citeyearpar{inline_attention}        & 16.7M & 5.0G & 80.2 (+0.4) \\
\rowcolor{verylightgray}
+ RPAttention   & 22.3M & 4.5G & \textbf{80.8 (+1.0)} \\
\midrule
PVT-T           & 13.2M & 1.9G & 75.1 \\
+ FLatten~\citeyearpar{flatten_transformer} & 12.2M & 2.0G & 77.8 (+2.7) \\
+ Agent~\citeyear{agent_attention}         & 11.6M & 2.0G & 78.4 (+3.3) \\
+ PolaFormer~\citeyearpar{polaformer}    & 12.0M & 2.0G & 78.8 (+3.7) \\
+ InLine~\citeyear{inline_attention}        & 12.0M & 2.0G & 78.2 (+3.1) \\
\rowcolor{verylightgray}
+ RPAttention   & 11.5M & 2.2G & \textbf{79.2 (+4.1)} \\
\midrule
PVT-S           & 24.5M & 3.8G & 79.8 \\
+ FLatten~\citeyearpar{flatten_transformer} & 21.7M & 4.0G & 81.7 (+1.9) \\
+ Agent~\citeyear{agent_attention}         & 20.6M & 4.0G & 82.2 (+2.4) \\
+ PolaFormer~\citeyearpar{polaformer}    & 21.0M & 4.1G & 81.9 (+2.1) \\
+ InLine~\citeyear{inline_attention}        & 21.6M & 3.9G & 82.0 (+2.2) \\
\rowcolor{verylightgray}
+ RPAttention   & 20.5M & 4.4G & \textbf{82.5 (+2.7)} \\
\midrule
PVT-M           & 44.2M & 6.7G & 81.2 \\
+ FLatten~\citeyearpar{flatten_transformer} & 37.2M & 7.0G & 83.0 (+1.8) \\
+ Agent~\citeyearpar{agent_attention}         & 35.9M & 7.0G & 83.4 (+2.2) \\
+ InLine~\citeyearpar{inline_attention}        & 37.6M & 6.9G & 83.2 (+2.0) \\
\rowcolor{verylightgray}
+ RPAttention   & 35.5M & 7.5G & \textbf{83.5 (+2.3)} \\
\midrule
PVT-L           & 61.4M & 9.8G  & 81.7 \\
+ FLatten~\citeyearpar{flatten_transformer} & 50.6M & 10.4G & 83.4 (+1.7) \\
+ Agent~\citeyearpar{agent_attention}       & 48.7M & 10.4G & \textbf{83.7 (+2.0)} \\
+ InLine~\citeyearpar{inline_attention}      & 50.2M & 10.2G & 83.6 (+1.9) \\
\rowcolor{verylightgray}
+ RPAttention   & 48.2M & 11.1G & 83.6 (+1.9) \\
\bottomrule
\end{tabular}
\end{minipage}\hfill
\begin{minipage}[t]{0.49\linewidth}
\centering
\begin{tabular}[t]{@{}lccc@{}}
\toprule
\textbf{Method} & \textbf{Params}
& \textbf{FLOPs} & \textbf{Top-1} \\
\midrule
PVTv2-B0        & 3.7M & 0.5G & 70.5 \\
+ FLatten~\citeyearpar{flatten_transformer} & 3.6M & 0.6G & 71.1 (+0.6) \\
+ MobiAtt~\citeyearpar{mobiatten} & 3.5M & 0.6G & 71.5 (+1.0) \\
+ PolaFormer~\citeyearpar{polaformer}    & 3.4M & 0.6G & 72.3 (+1.8) \\
\rowcolor{verylightgray}
+ RPAttention   & 3.2M & 0.7G & \textbf{73.1 (+2.6)} \\
\midrule
PVTv2-B1        & 13.1M & 2.1G & 78.7 \\
+ FLatten~\citeyearpar{flatten_transformer} & 12.9M & 2.2G & 79.5 (+0.8) \\
+ PolaFormer~\citeyearpar{polaformer}    & 13.0M & 2.2G & 80.2 (+1.5) \\
\rowcolor{verylightgray}
+ RPAttention   & 12.3M & 2.4G & \textbf{80.5 (+1.8)} \\
\midrule
PVTv2-B2        & 25.4M & 4.0G & 82.0 \\
+ FLatten~\citeyearpar{flatten_transformer} & 22.6M & 4.3G & 82.5 (+0.5) \\
+ MobiAtt~\citeyearpar{mobiatten} & 21.1M & 3.8G & \textbf{82.6 (+0.6)} \\
\rowcolor{verylightgray}
+ RPAttention   & 21.3M & 4.6G & \textbf{82.6 (+0.6)} \\
\midrule
Swin-T          & 29.0M & 4.5G & 81.3 \\
+ FLatten~\citeyearpar{flatten_transformer} & 29.0M & 4.5G & 82.1 (+0.8) \\
+ Agent~\citeyearpar{agent_attention}         & 29.0M & 4.5G & 82.6 (+1.3) \\
+ PolaFormer~\citeyearpar{polaformer}    & 29.0M & 4.5G & 82.6 (+1.3) \\
+ InLine~\citeyearpar{inline_attention}        & 30.0M & 4.5G & 82.4 (+1.1) \\
+ CA~\citeyearpar{ca_learners} & 28.0M & 4.6G & 82.2 (+0.9) \\
\rowcolor{verylightgray}
+ RPAttention   & 28.4M & 4.9G & \textbf{82.7 (+1.4)} \\
\midrule
Swin-S          & 50M & 8.7G & 83.0 \\
+ FLatten~\citeyearpar{flatten_transformer} & 51M & 8.7G & 83.5 (+0.5) \\
+ Agent~\citeyearpar{agent_attention}         & 50M & 8.7G & 83.7 (+0.7) \\
+ PolaFormer~\citeyearpar{polaformer}    & 50M & 8.7G & 83.6 (+0.6) \\
+ InLine~\citeyearpar{inline_attention}        & 50M & 8.7G & 83.6 (+0.6) \\
+ CA~\citeyearpar{ca_learners} & 50M & 8.8G & 83.6 (+0.6) \\
\rowcolor{verylightgray}
+ RPAttention   & 50M & 9.3G & \textbf{83.8 (+0.8)} \\
\bottomrule
\end{tabular}
\end{minipage}

\endgroup
\end{table}

\subsection{Dense Prediction}
\label{app:dense_comparisons}
Table~\ref{tab:dense_method_comparisons} compares RPAttention with
other efficient attention methods on COCO and ADE20K.
RPAttention consistently ranks among the top two on the reported
AP, mIoU, and mAcc metrics within each backbone group.
On COCO, it achieves the best or tied-best box AP and the best
mask AP across all three PVT variants.
On ADE20K, it achieves the highest mIoU in most backbone groups.
These results highlight its strong performance across dense
prediction tasks.
\begin{table}[H]
\centering
\caption{Comparison with other efficient-attention methods on
COCO and ADE20K. COCO results use Mask R-CNN with the
$1\times$ schedule. Unavailable entries are denoted by ``--''.
Bold and underlined values indicate the best and second-best
distinct scores within each backbone group, respectively.}
\label{tab:dense_method_comparisons}
\label{tab:coco_method_comparisons}
\label{tab:ade20k_method_comparisons}

\begingroup
\setlength{\tabcolsep}{0.8mm}
\setlength{\aboverulesep}{0pt}
\setlength{\belowrulesep}{0pt}
\renewcommand{\arraystretch}{1.12}

\resizebox{\linewidth}{!}{%
\begin{tabular}{@{}lcccccccccc@{}}
\toprule
\multirow{2}{*}{\textbf{Method}}
& \multicolumn{7}{c}{\textbf{Detection \& Instance Seg.}}
& \multicolumn{3}{c}{\textbf{Semantic Seg.}} \\
\cmidrule(lr){2-8}
\cmidrule(l){9-11}
& \textbf{FLOPs}
& $\mathbf{AP}^{b}$
& $\mathbf{AP}^{b}_{50}$
& $\mathbf{AP}^{b}_{75}$
& $\mathbf{AP}^{m}$
& $\mathbf{AP}^{m}_{50}$
& $\mathbf{AP}^{m}_{75}$
& \textbf{FLOPs}
& \textbf{mIoU}
& \textbf{mAcc} \\
\midrule

PVT-T
& 240G & 36.7 & 59.2 & 39.3 & 35.1 & 56.7 & 37.3
& 158G & 36.57 & 46.72 \\
+ FLatten~\citeyearpar{flatten_transformer}
& 244G & \underline{38.2} & 61.6 & 41.9
& 37.0 & 57.6 & 39.0
& 169G & 37.21 & 48.95 \\
+ PolaFormer~\citeyearpar{polaformer}
& 216G & \textbf{40.4} & \underline{62.4} & \textbf{43.9}
& \underline{37.4} & \underline{59.4} & \underline{40.3}
& 133G & 38.3 & -- \\
+ InLine~\citeyearpar{inline_attention}
& -- & -- & -- & -- & -- & -- & --
& 127G & \underline{39.16} & \underline{50.63} \\
\rowcolor{verylightgray}
+ RPAttention
& 215G & \textbf{40.4} & \textbf{62.8} & \underline{43.8}
& \textbf{37.7} & \textbf{59.9} & \textbf{40.4}
& 128G & \textbf{40.00} & \textbf{53.54} \\
\midrule

PVT-S
& 305G & 40.4 & 62.9 & 43.8 & 37.8 & 60.1 & 40.3
& 225G & 41.95 & 53.02 \\
+ Agent~\citeyearpar{agent_attention}
& -- & -- & -- & -- & -- & -- & --
& 211G & \textbf{44.18} & \textbf{56.17} \\
+ PolaFormer~\citeyearpar{polaformer}
& 262G & \underline{43.9} & 66.1 & \underline{47.9}
& \underline{40.2} & \underline{63.1} & 43.0
& 179G & 41.0 & -- \\
+ InLine~\citeyearpar{inline_attention}
& 250G & 43.4 & \underline{66.4} & 47.1
& 40.1 & \underline{63.1} & \underline{43.3}
& 168G & 42.93 & 54.58 \\
\rowcolor{verylightgray}
+ RPAttention
& 259G & \textbf{44.1} & \textbf{66.6} & \textbf{48.5}
& \textbf{40.8} & \textbf{63.7} & \textbf{44.3}
& 177G & \underline{43.25} & \underline{56.04} \\
\midrule

PVT-M
& 392G & 42.0 & 64.4 & 45.6 & 39.0 & 61.6 & 42.1
& 315G & 42.91 & 53.80 \\
+ Agent~\citeyearpar{agent_attention}
& -- & -- & -- & -- & -- & -- & --
& 321G & 44.30 & 56.42 \\
+ InLine~\citeyearpar{inline_attention}
& 310G & \underline{44.0} & \underline{66.4} & \underline{48.0}
& \underline{40.3} & \underline{63.4} & \underline{43.5}
& 229G & \underline{44.59} & \underline{57.20} \\
\rowcolor{verylightgray}
+ RPAttention
& 322G & \textbf{45.9} & \textbf{68.2} & \textbf{50.4}
& \textbf{41.9} & \textbf{65.4} & \textbf{45.3}
& 241G & \textbf{45.16} & \textbf{58.13} \\
\midrule

Swin-T
& 267G & 43.7 & 66.6 & 47.7
& \underline{39.8} & 63.3 & 42.7
& 945G & 44.51 & 55.61 \\
+ FLatten~\citeyearpar{flatten_transformer}
& -- & -- & -- & -- & -- & -- & --
& 946G & 44.82 & 57.01 \\
+ Agent~\citeyearpar{agent_attention}
& 261G & 41.4 & 63.5 & 45.2 & 38.9 & 60.2 & 40.2
& 954G & \underline{46.68} & \underline{58.53} \\
+ PolaFormer~\citeyearpar{polaformer}
& 272G & \textbf{44.8} & \textbf{67.6} & \textbf{49.1}
& \textbf{40.5} & \underline{64.1} & \textbf{43.5}
& 946G & 45.8 & -- \\
+ InLine~\citeyearpar{inline_attention}
& -- & -- & -- & -- & -- & -- & --
& 941G & 45.57 & 57.60 \\
\rowcolor{verylightgray}
+ RPAttention
& 271G & \underline{44.5} & \underline{67.3} & \underline{48.4}
& \textbf{40.5} & \textbf{64.2} & \underline{43.3}
& 949G & \textbf{46.97} & \textbf{59.17} \\
\midrule

Swin-S
& 358G & 45.7 & 67.9 & \underline{50.4}
& 41.1 & 64.9 & 44.2
& 1038G & 47.64 & 58.78 \\
+ FLatten~\citeyearpar{flatten_transformer}
& -- & -- & -- & -- & -- & -- & --
& 1038G & \underline{48.14} & 59.31 \\
+ Agent~\citeyearpar{agent_attention}
& 364G & \textbf{47.2} & \textbf{69.6} & \textbf{52.3}
& \textbf{42.7} & \textbf{66.6} & \textbf{45.8}
& 1043G & 48.08 & \underline{59.78} \\
\rowcolor{verylightgray}
+ RPAttention
& 360G & \underline{46.3} & \underline{69.0} & \underline{50.4}
& \underline{41.8} & \underline{65.7} & \underline{45.1}
& 1042G & \textbf{48.48} & \textbf{60.12} \\

\bottomrule
\end{tabular}%
}
\endgroup
\end{table}
\clearpage

\section{Additional Analyses and Visualizations}

\subsection{Cross-Seed Routing Stability}
\label{app:representative_behavior}
We examine whether RPAttention learns consistent spatial routing
across independent training runs.
Two RPAttention-PVTv2-B1 models are trained with identical settings
using random seeds 123 and 456.
To compare their routing independently of representative ordering,
we use the composed query-to-space response in
Eq.~\ref{eq:composed_response}.

\refstepcounter{equation}
\label{eq:composed_response}
$$
R_{q,\mathrm{full}}
=
\operatorname{Normalize}_{\mathrm{space}}
\left[
\frac{1}{H}
\sum_{h=1}^{H}
D_{q,:}^{(h)}
\left(I+S^{(h)}\right)
\hat{A}^{(h)\top}
\right].
\eqno(\theequation)
$$

where $H$ is the number of attention heads, $I$ is the identity matrix,
and $\hat{A}$, $S$, and $D$ denote the normalized Gather assignment,
Interact attention, and Distribute attention, respectively.
The operator $\operatorname{Normalize}_{\mathrm{space}}$ normalizes
each response over spatial locations.

The composed response in Eq.~\ref{eq:composed_response} is invariant
to the ordering of representatives. Under an arbitrary representative
permutation $P$, the corresponding quantities transform as
$\hat{A}'=\hat{A}P^\top$, $D'=DP^\top$, and $S'=PSP^\top$.
Consequently,

\refstepcounter{equation}
\label{eq:permutation_invariance}
$$
\begin{aligned}
D'
\left(I+S'\right)
\hat{A}'^{\top}
&=
DP^\top
\left(I+PSP^\top\right)
P\hat{A}^{\top}
\\
&=
D\left(I+S\right)\hat{A}^{\top}.
\end{aligned}
\eqno(\theequation)
$$

Equation~\ref{eq:permutation_invariance} shows that the representative
permutations cancel in the composed spatial response.

\begin{table}[H]
\centering
\caption{
Cross-seed stability of composed query-to-space responses for matched
image--query pairs on 1,000 COCO-val images. Metrics are averaged over
queries, and brackets report 95\% confidence intervals obtained by
bootstrap resampling images.
}
\label{tab:seed}
\small
\setlength{\tabcolsep}{5pt}
\renewcommand{\arraystretch}{1.08}
\begin{tabular}{lccc}
\toprule
\textbf{Stage}
& \textbf{Cosine $\uparrow$}
& \textbf{JS $\downarrow$}
& \textbf{Top-20\% IoU $\uparrow$} \\
\midrule
Stage 3
& 0.9491 [0.9484, 0.9497]
& 0.0125 [0.0123, 0.0126]
& 0.5421 [0.5382, 0.5460] \\

Stage 4
& 0.9610 [0.9601, 0.9619]
& 0.0098 [0.0096, 0.0100]
& 0.4860 [0.4795, 0.4926] \\
\bottomrule
\end{tabular}
\end{table}

On 1,000 COCO-val images evaluated at the same nine query locations
in the last blocks of Stages 3 and 4, the independently trained models
exhibit high agreement in their composed response distributions, with
cosine similarities above 0.94 and low JS divergences
(Table~\ref{tab:seed}). The Top-20\% IoU values are more moderate,
indicating that the strongest-response regions are similar but not
identical across training runs.

As an additional control, we evaluate the direct Gather--Distribute
response at Stage 3 by replacing $(I+S^{(h)})$ with $I$ in
Eq.~\ref{eq:composed_response}.
This excludes representative interaction from the response composition
while preserving invariance to representative ordering.
The cross-seed cosine similarity is 0.9107 for the same image and query,
compared with 0.6470 for the same image with different queries
and 0.6154 for different images at the same query coordinates.
These results support input- and query-dependent routing consistency
that cannot be attributed solely to mixing through Interact.

\subsection{Foreground Alignment Analysis}
\label{app:foreground}

On 1{,}000 class-stratified ImageNet validation images with localization annotations, we analyze Agent Attention--PVT-S and RPAttention--PVT-S using identical preprocessing. 
We measure: (1) \emph{foreground concentration} (attention mass in the ground-truth region divided by its area fraction; uniform attention scores 1); 
(2) \emph{pointing accuracy} (whether the response maximum lies in a ground-truth box); and (3) \emph{map redundancy} (mean pairwise cosine similarity among representative maps). 
Each entry in Table~\ref{tab:foreground} reports Agent Attention / RPAttention. 
RPAttention achieves higher foreground concentration and pointing
accuracy, together with lower map redundancy, at every stage
(Table~\ref{tab:foreground}).
All paired differences have image-level bootstrap 95\% confidence
intervals excluding zero.
At Stage 4, RPAttention also exhibits a higher normalized effective
rank (0.247 vs.\ 0.193) and greater adjacent-layer consistency
(0.768 vs.\ 0.720), while maintaining comparable flip consistency
(0.897 vs.\ 0.901).
Together, these observations indicate that RPAttention forms
more distinct representative maps with stronger foreground alignment
and consistent responses across layers and horizontal flips.

\begin{table}[h]
\centering
\caption{Foreground alignment of intermediate tokens (Agent Attention / RPAttention on PVT-S, ImageNet-1K validation).}
\label{tab:foreground}
\begingroup
\setlength{\tabcolsep}{6pt}
\begin{tabular}{lccc}
\toprule
\textbf{Stage} & \textbf{Foreground concentration $\uparrow$} & \textbf{Pointing accuracy $\uparrow$} & \textbf{Map redundancy $\downarrow$} \\
\midrule
1 & 0.961 / \textbf{1.115} & 0.568 / \textbf{0.649} & 0.378 / \textbf{0.089} \\
2 & 0.999 / \textbf{1.146} & 0.571 / \textbf{0.660} & 0.395 / \textbf{0.242} \\
3 & 1.226 / \textbf{1.325} & 0.643 / \textbf{0.662} & 0.260 / \textbf{0.230} \\
4 & 0.947 / \textbf{1.039} & 0.491 / \textbf{0.529} & 0.807 / \textbf{0.532} \\
\bottomrule
\end{tabular}
\endgroup
\end{table}

\subsection{Additional Query-Response Visualizations}
\label{app:qualitative}
\begin{figure}[H]
  \centering
  \includegraphics[width=\linewidth]{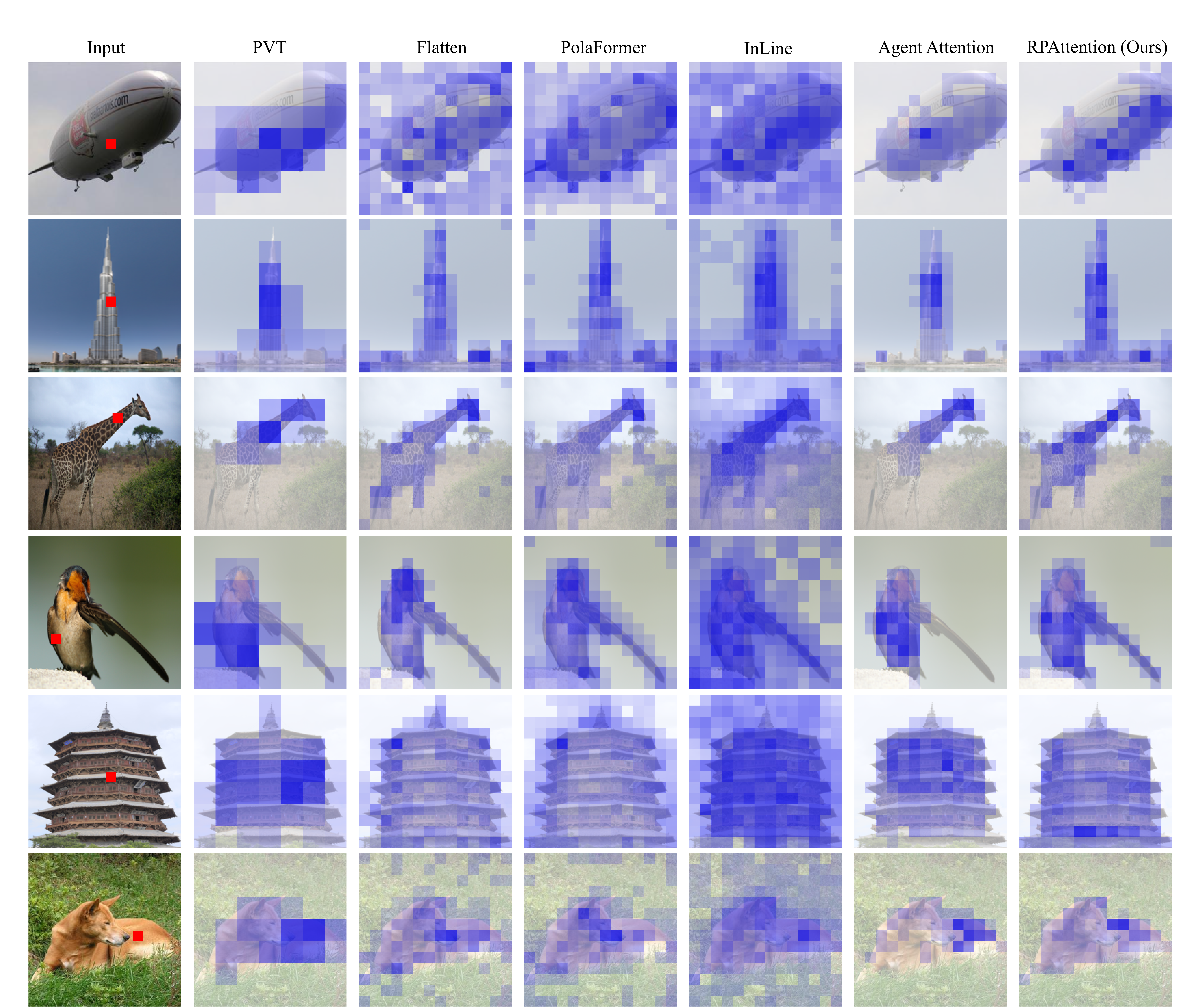}
  \caption{Additional query-response visualizations. For selected query locations, 
  RPAttention produces response maps that align with semantically related image regions, 
  illustrating how representatives mediate global information exchange beyond fixed spatial layouts.}
  \label{fig:query_response}
\end{figure}

Figure~\ref{fig:query_response} provides additional examples of
how RPAttention connects spatial queries with image content,
complementing the comparisons in Figure~\ref{fig:analysis}.

\begingroup

\subsection{Representative Construction for Classification and Dense Prediction}

To examine the effect of representative construction, we compare
RPAttention with a pooled-query Gather variant on DeiT-Tiny.
The Interact, Distribute, and DWC components are retained in both
variants. We evaluate both models on ImageNet-1K classification and
ADE20K semantic segmentation. For segmentation, each model is initialized
with its corresponding ImageNet-pretrained weights and trained using the
same Semantic FPN configuration for 40k iterations.

\begin{table}[htbp]
\centering
\caption{
Comparison of representative construction on DeiT-Tiny for ImageNet-1K
classification and ADE20K semantic segmentation.
}
\label{tab:gather_classification_segmentation}
\begin{tabular}{lcc}
\toprule
Method & ImageNet Top-1 (\%) $\uparrow$ & ADE20K mIoU (\%) $\uparrow$ \\
\midrule
RPAttention with pooled-query Gather & 75.2 & 35.53 \\
RPAttention & \textbf{75.4} & \textbf{37.88} \\
\bottomrule
\end{tabular}
\end{table}

As shown in Table~\ref{tab:gather_classification_segmentation},
the two variants achieve similar classification accuracy, while
RPAttention improves segmentation mIoU by 2.35 percentage points in this
setting. Classification focuses on recognizing the content of an image
as a whole, whereas segmentation requires predictions at individual
spatial locations. This comparison suggests that feature-dependent representative
construction can benefit dense prediction, even when the difference
in classification accuracy is small.

\endgroup

\begingroup

\subsection{Representative Utilization}

To examine whether token assignments concentrate on only a few
representatives, we analyze a trained RPAttention-PVT-Tiny model on
5,000 ImageNet validation images. 

For each image and attention head, we average the weights assigned to
each representative over spatial tokens, obtaining a distribution
$p_m$. We compute the effective number of representatives as
$N_{\mathrm{eff}}=\exp(-\sum_m p_m\log p_m)$, and report the mean
across images and heads. For Gather, we use the assignment weights
before normalization across tokens; for Distribute, we use the
query-to-representative attention weights.

\begin{table}[htbp]
\centering
\caption{
Effective number of representatives in the final block of
RPAttention-PVT-Tiny.
}
\label{tab:representative_utilization}
\begin{tabular}{lcc}
\toprule
Stage & Number of representatives & Effective number \\
\midrule
Gather     & 49 & 20.24 \\
Distribute & 49 & 37.02 \\
\bottomrule
\end{tabular}
\end{table}

As shown in Table~\ref{tab:representative_utilization}, Gather
assignments are distributed across multiple representatives, with an
effective count of 20.24 out of 49. The effective count increases to
37.02 for Distribute, indicating a more even readout distribution
across representatives when averaged over spatial positions.

The two stages compute their weights separately. Gather assigns tokens
according to their compatibility with learned directions, whereas
Distribute matches spatial queries to representative keys. The readout
distribution therefore need not mirror the Gather assignments. In this
setting, the broader aggregate readout allows multiple updated
representatives to contribute to the spatial outputs, supporting the
combination of compact aggregation and query-conditioned retrieval.
These distributions emerge without an additional load-balancing loss.

\endgroup

\end{document}